\documentclass{article}

\usepackage{PRIMEarxiv}

\usepackage[utf8]{inputenc}
\usepackage[T1]{fontenc}
\usepackage{url}
\usepackage{booktabs}
\usepackage{amsfonts}
\usepackage{nicefrac}
\usepackage{microtype}
\usepackage{fancyhdr}
\usepackage{graphicx}
\graphicspath{{images/}}
\usepackage[dvipsnames,table]{xcolor}
\usepackage{amsmath}
\usepackage{amssymb}
\usepackage{mathtools}
\usepackage{amsthm}
\usepackage{xspace}
\usepackage{wrapfig}
\usepackage{multirow}
\usepackage{makecell}
\usepackage{colortbl}
\usepackage{subcaption}
\usepackage{caption}
\usepackage{tcolorbox}
\usepackage{algorithm}
\usepackage{algorithmic}
\usepackage[breaklinks,colorlinks]{hyperref}
\usepackage[capitalize,noabbrev]{cleveref}

\title{Look Less, Reason More: Block-wise Attention Skipping for Efficient Multimodal LLMs}

\author{
  Jie Ma~~ Zhike Qiu~~ Jiayi Ji~~ Xiaoshuai Sun~~ Rongrong Ji \\
 \normalsize Xiamen University\\
  \tt\small {jiema100@stu.xmu.edu.cn}, \texttt{zhikeqiu@outlook.com} \\
  \tt\small {jjyxmu@gmail.com}, \tt\small{xxsun@xmu.edu.cn},
  \tt\small {rrji@xmu.edu.cn} \\
}

\begin{document}
\maketitle

\begin{abstract}
Multimodal Large Language Models (MLLMs) face a significant inference bottleneck due to the quadratic computational cost of self-attention over long visual token sequences. However, we identify a critical inefficiency in current architectures: \textit{\textbf{Visual Attention Saturation}}. Our analysis reveals that visual tokens rapidly establish their spatial structure and intra-modal relationships in early layers, rendering visual-to-visual self-attention in deeper layers computationally redundant. Conversely, Feed-Forward Networks in these layers remain essential for projecting visual features into the evolving textual semantic space. Leveraging this insight, we present Visual-Skip (V-Skip), a training-free inference paradigm that decouples spatial interaction from semantic evolution. Rather than discarding tokens, V-Skip imposes block-wise structured sparsity by selectively bypassing saturated visual self-attention modules. Furthermore, recognizing that varying downstream tasks demand distinct reasoning depths, V-Skip employs a lightweight, few-shot calibration to dynamically route the task-optimal sparsity path. Extensive experiments demonstrate that V-Skip effectively bypasses redundant vision attention to achieve block-wise sparsity, maintaining a 94.16\% to 100.31\% performance retention across diverse MLLMs. Ultimately, we prove that to reason more effectively, models do not need to discard what they see --- they simply need to ``look less'' at the right depth.
\end{abstract}

\section{Introduction}
\label{sec:intro}
Multimodal Large Language Models (MLLMs) \cite{llava1,llava,llavanext,llavaOneVision,deepseekai2025deepseekv3technicalreport,qwenvl25} have demonstrated remarkable capabilities in visual understanding and reasoning. To capture fine-grained details in complex scenes, recent models increasingly adopt high-resolution image input \cite{llavanext,qwen3technicalreport}, representing visual content as long sequences of visual tokens (e.g., expanding from 576 to over 2800 tokens). While this higher resolution enhances perception, it introduces a severe computational bottleneck. The self-attention mechanism in Transformers exhibits quadratic complexity with respect to the sequence length. Consequently, as the number of visual tokens grows, the computational cost and memory consumption during the prefill phase surge disproportionately, resulting in prohibitive latency for real-time applications.

\begin{figure}[tb]
  \centering
  \includegraphics[width=\linewidth]{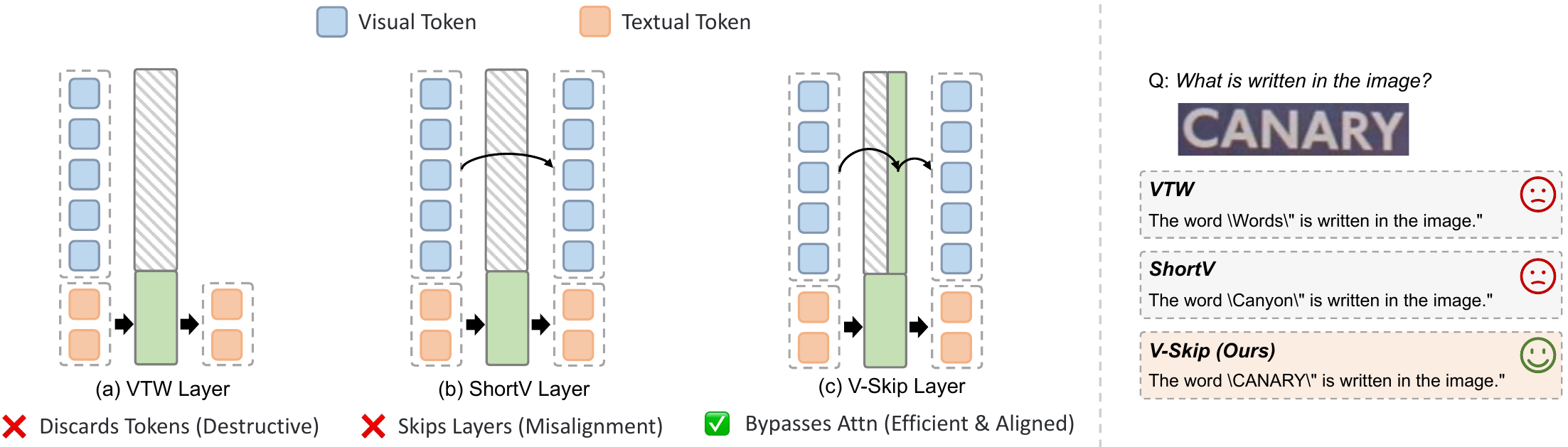}
    \caption{Conceptual and qualitative comparison of visual acceleration paradigms. (Left) Existing methods operate destructively: VTW~\cite{vtw} drops tokens (losing context), and ShortV~\cite{shortv} skips entire layers (disrupting semantic evolution). Our V-Skip uniquely decouples the block, bypassing redundant spatial attention while keeping FFNs active for continuous semantic alignments.
    (Right) A real-world OCR example illustrating the consequences of these paradigms. VTW loses fine-grained visual details (outputting generic ``Words''). ShortV suffers from feature misalignment due to missing FFNs, leading to semantic hallucination (``Canyon''). V-Skip successfully maintains full visual and semantic integrity (``CANARY'').
  }
  \label{fig:fig1}
\end{figure}

To mitigate this inefficiency, recent research has focused on reducing the number of visual tokens. Prominent methods, such as Token Pruning \cite{Fastv,PyramidDrop,sparsevlm,vispruner} and Token Merging \cite{tome}, accelerate inference by aggressively identifying and discarding ``redundant'' tokens. As illustrated in \cref{fig:fig1} (Left)(a,b), more aggressive approaches \cite{vtw,shortv} extend this paradigm by progressively terminating visual computation in intermediate layers or freezing visual representations by skipping entire transformer blocks. While effective in reducing FLOPs, these approaches are inherently destructive. Specifically, by permanently removing tokens or skipping layer-wise feature updates, they irreversibly compromise visual integrity or disrupt feature evolution. This often leads to a loss of fine-grained details and an increased risk of object hallucination as intuitively demonstrated by the OCR failure case in \cref{fig:fig1} (Right), where the model fails to ground its reasoning in the actual image content. This creates a challenging trade-off: \textit{current acceleration methods sacrifice visual integrity for inference speed.}

To break this trade-off, we move beyond the coarse-grained paradigm of indiscriminately dropping tokens or freezing entire layers. Instead, we ask: \textit{Are all operations within a deep transformer block equally redundant for visual tokens?} 
Through a layer-wise analysis of attention patterns, we identify a critical phenomenon termed \textbf{Visual Attention Saturation}. We observe that visual processing in MLLMs undergoes a functional evolution across layers. In early layers, visual tokens require dense spatial interaction (Self-Attention) to aggregate local features into coherent objects. However, in deeper layers, this spatial structure stabilizes, and the intra-modal attention maps become static. At this stage, the primary role of visual tokens shifts from ``spatial construction'' to ``semantic reasoning'', serving as keys for the Language Model to query~\cite{multimodelProcess,visualrepresentationsmap}. Crucially, while the quadratic self-attention computation becomes redundant in these deep layers, the token-wise transformation via Feed-Forward Networks (FFNs) remains essential for aligning visual features with the evolving semantic space of the Large Language Model.

Guided by this insight, we introduce Visual-Skip (V-Skip), a simple yet effective training-free acceleration strategy for MLLMs, as illustrated in \cref{fig:fig1} (Left)(c). Unlike pruning methods that drop tokens, V-Skip preserves the complete sequence of visual tokens throughout the network. Instead, it introduces block-wise structured sparsity by selectively bypassing the visual self-attention computation in deep, saturated layers while fully retaining the FFN transformations. Furthermore, recognizing that diverse downstream tasks require varying depths of spatial reasoning, V-Skip employs a highly efficient few-shot calibration mechanism to dynamically identify the task-optimal sparsity path. This approach effectively decouples spatial interaction from semantic alignment, allowing the model to ``Look Less'' (eliminating quadratic redundancy) while continuing to ``Reason More'' (maintaining semantic depth). By avoiding the structural damage caused by token dropping, V-Skip ensures that the model retains full access to visual context for complex reasoning tasks.

Our contributions can be summarized as follows:
\begin{itemize}
    \item We uncover the phenomenon of Visual Attention Saturation in MLLMs, revealing that deep-layer visual self-attention becomes computationally redundant, whereas FFNs remain critical for layer-wise semantic alignment.
    \item  We introduce V-Skip, a training-free paradigm that decouples spatial interaction from semantic evolution. It achieves block-wise sparsity by bypassing saturated attention modules without retraining.
    \item Extensive experiments validate the superiority of our paradigm. V-Skip achieves an impressive 98.42\% to 100.31\% performance retention across the LLaVA series, alongside a robust 94.16\% on the Qwen architecture. Notably, by preserving complete visual context, it overcomes the structural damage of traditional pruning methods, significantly mitigating object hallucinations in complex reasoning tasks.

\end{itemize}

\section{Related Work}

\noindent\textbf{MLLMs.}
The rapid evolution of Multimodal Large Language Models (MLLMs) has been driven by the integration of powerful visual encoders with Large Language Models (LLMs) \cite{llava1,llava,llavaOneVision,llavanext,qwenvl25,deepseekai2025deepseekv3technicalreport}. While early models operated on low-resolution inputs (e.g., $224^2$ or $336^2$), recent advancements like LLaVA-NeXT \cite{llavanext}, GPT \cite{gpt4}, and Gemini \cite{Gemini} have shifted towards high-resolution visual processing to capture fine-grained details and text in images. For instance, LLaVA-NeXT adopts an ``AnyRes'' strategy, dynamically tiling images into multiple patches, which can increase the visual token sequence length from 576 to over 2800 tokens.
However, this resolution scaling comes at a steep computational cost. Since the self-attention mechanism in Transformers scales quadratically with sequence length \cite{vit}, the latency and memory usage grow prohibitively in high-resolution settings. This creates an urgent need for efficient inference strategies that can handle long visual sequences without compromising the model's perceptual capabilities, driving the exploration of structural redundancy.

\noindent\textbf{Efficiency via Redundancy Reduction.}
To mitigate the computational overhead of high-resolution visual sequences, prior research has primarily divided into token reduction and structural pruning paradigms. Token reduction strategies \cite{Fastv,PyramidDrop,sparsevlm,vispruner,vtw} accelerate inference by progressively discarding visual token computation. However, these methods operate on a destructive premise: the permanent exclusion or premature withdrawal of tokens interrupts continuous visual context, often exacerbating object hallucination due to compromised visual representations \cite{visionzip}.
Conversely, structural approaches \cite{shortv,vtw} target architectural redundancy by skipping entire transformer blocks or ``freezing'' visual representations. Crucially, simply bypassing the Feed-Forward Networks neglects the necessity of layer-wise semantic evolution \cite{ffn_layers}, resulting in feature misalignment between the frozen visual tokens and the increasingly abstract textual manifold \cite{semanticevolution}. Diverging from these paradigms, we identify a modality-specific functional decoupling: while layer-wise spatial interaction (Attention) saturates, semantic alignment (FFN) remains indispensable. Guided by this insight, our V-Skip mechanism selectively bypasses visual attention to eliminate quadratic redundancy, avoiding both the information loss inherent in pruning and the semantic misalignment caused by layer freezing.
\begin{figure}[tb]
  \centering
  \includegraphics[width=0.98\linewidth]{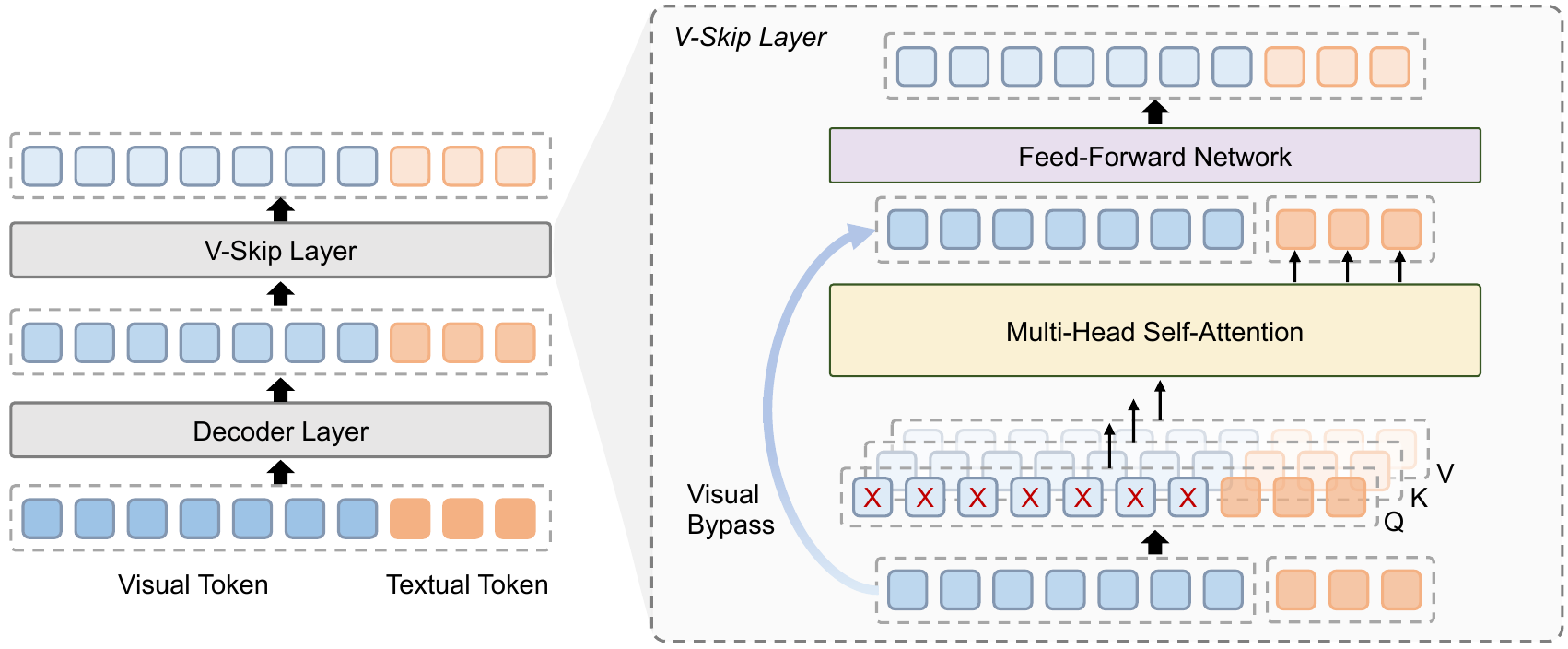}
  \caption{Illustration of the V-Skip.
In identified visual attention saturated layers, we decouple the computational path of visual and textual tokens. Visual tokens (blue) utilize a Visual Bypass to skip the self-attention calculation, denoted by the red crosses. Textual tokens (orange) maintain full attention to both visual and textual contexts for robust reasoning. Unlike layer-dropping methods, V-Skip processes all tokens through the Feed-Forward Network, ensuring continuous layer-wise semantic evolution and alignment. Best viewed in color.
  }
  \label{fig:arch}
\end{figure}

\section{Methodology}
\label{sec:method}
In this section, we introduce the proposed V-Skip paradigm and describe its design in detail. We begin by formalizing the computational bottleneck in MLLMs in \cref{sec:Formulation}, then present the Visual Information Gain metric in \cref{subsec:saturation} for identifying saturated blocks, and finally \cref{subsec:vskip} details the block-wise attention skipping mechanism, as illustrated in \cref{fig:arch}.

\subsection{Problem Formulation}
\label{sec:Formulation}

We consider a Multimodal Large Language Model (MLLM) parameterized by $\theta$, which processes a multimodal sequence consisting of visual inputs $\mathbf{I}$ and textual instructions $\mathbf{X}_{t}$. The visual inputs are encoded into a sequence of visual embeddings $\mathbf{X}_{v} \in \mathbb{R}^{N_v \times d}$ via a vision encoder and a projector, where $N_v$ denotes the number of visual tokens. The textual inputs are tokenized into $\mathbf{X}_{t} \in \mathbb{R}^{N_t \times d}$. The concatenated sequence $\mathbf{X} = [\mathbf{X}_{v}; \mathbf{X}_{t}] \in \mathbb{R}^{N \times d}$ serves as the input to the LLM backbone, where $N = N_v + N_t$.

The LLM backbone consists of $L$ stacked Transformer decoder layers. For a specific layer $l$, the hidden state $\mathbf{H}^l \in \mathbb{R}^{N \times d}$ is updated via Multi-Head Self-Attention (MHSA) and a Feed-Forward Network (FFN):
\begin{align}
    \mathbf{H}'^l &= \text{MHSA}(\text{LN}(\mathbf{H}^l)) + \mathbf{H}^l, \label{eq:mhsa} \\
    \mathbf{H}^{l+1} &= \text{FFN}(\text{LN}(\mathbf{H}'^l)) + \mathbf{H}'^l, \label{eq:ffn}
\end{align}
where $\text{LN}(\cdot)$ denotes Layer Normalization.

\noindent\textbf{Computational Bottleneck.} The complexity of the MHSA operation in Eq.~\eqref{eq:mhsa} is $\mathcal{O}(N^2)$. In high-resolution settings (e.g., LLaVA-NeXT), $N_v \gg N_t$. Consequently, the intra-modal attention among visual tokens, which constitutes a complexity of $\mathcal{O}(N_v^2)$, dominates the prefill latency.

\subsection{Visual Attention Saturation: A Block-wise Analysis}
\label{subsec:saturation}

Existing acceleration methods typically assume redundancy in the quantity of visual tokens. In contrast, we investigate the redundancy in the computational structure across network depth. The visual representation learning in MLLMs undergoes a functional phase transition~\cite{multimodelProcess}: from spatial aggregation in early layers to semantic alignment in deep layers.

To rigorously quantify this transition, we propose the \textbf{Visual Information Gain (VIG)} metric based on prediction sensitivity. Rather than measuring intermediate feature norms, we evaluate the impact of bypassing visual attention at a specific layer $l$ on the model's final probability distribution.

Let $P(\mathbf{y} | \mathbf{X}; \theta)$ denote the predictive distribution of the full model over the vocabulary given input $\mathbf{X}$. Let $P_{\neg l}(\mathbf{y} | \mathbf{X}; \theta)$ denote the distribution obtained when the intra-modal visual self-attention at layer $l$ is bypassed (i.e., skipping the $\mathcal{O}(N_v^2)$ mixing), while keeping the FFN active. Let $\mathcal{D}_k$ denote a specific downstream task dataset (e.g., OCR, VQA, or Chart Understanding). We define the Task-Specific VIG for layer $l$ under task $k$, denoted as $\mathcal{G}_k(l)$, as the expected Kullback-Leibler (KL) divergence computed over a calibration set randomly sampled from $\mathcal{D}_k$:

\begin{equation}
\mathcal{G}_k(l) = \mathbb{E}_{X \sim \mathcal{D}_k} \left[ D_{KL}(P(y|X) || P_{\neg l}(y|X)) \right]
\label{eq:vig_kl}
\end{equation}

To capture these task-specific dynamics with minimal overhead, we perform a lightweight few-shot calibration by randomly sampling merely 20 image-text pairs from $\mathcal{D}_k$ to compute this expectation. This highly efficient sampling strategy guarantees that the measured saturation pattern accurately reflects the unique reasoning pathway required by the specific task, avoiding the bias of a unified mask (e.g., ShortV) while imposing negligible pre-computation burdens.

\begin{figure}[tb]
  \centering
  \includegraphics[width=\linewidth, trim=0 0 0 0, clip]{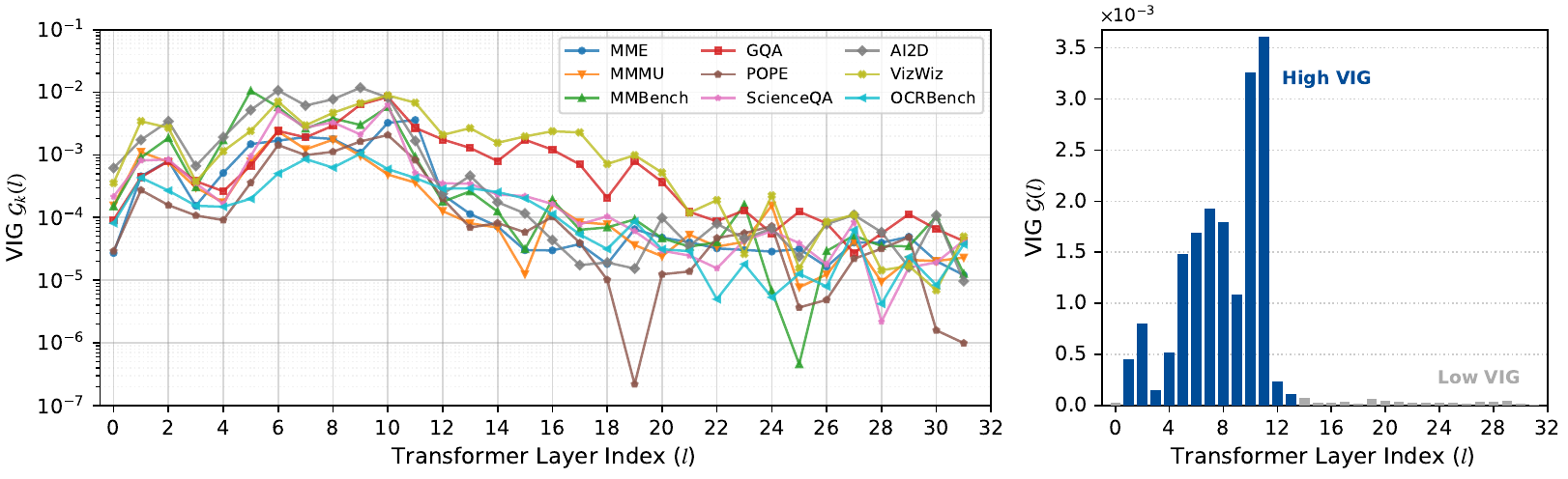}
  \caption{Empirical analysis of Visual Attention Saturation. (Left) Task-specific VIG profiles $\mathcal{G}_k(l)$ across nine representative multimodal benchmarks. While all tasks generally exhibit a transition toward saturation in deep layers, their specific trajectories and attention demands vary significantly, highlighting the necessity of task-aware calibration. (Right) A detailed block-wise VIG bar chart using the MME dataset as a case study. A stark phase transition is observed: early layers (blue bars) actively aggregate spatial information, whereas deeper layers (grey bars) reach attention saturation, rendering further intra-modal queries computationally redundant.}
  \label{fig:vig_analysis}
\end{figure}

\noindent\textbf{Empirical Observation of Saturation}. To validate visual attention saturation, we visualize the computed VIG across the 32 transformer layers with LLaVA-1.5-7B. As explicitly shown in \cref{fig:vig_analysis} (Right) using the MME benchmark as a case study, $\mathcal{G}(l)$ reveals a stark and undeniable phase transition. Early and intermediate layers (primarily indices 1--12) exhibit exceptionally high information gain, confirming their critical role in active spatial aggregation and object formulation. Conversely, the VIG plunges drastically and flattens below the adaptive threshold in deeper layers, empirically verifying the Visual Attention Saturation phenomenon.

Crucially, our expanded analysis across multiple benchmarks in \cref{fig:vig_analysis} (Left) uncovers a deeper architectural insight: saturation profiles are inherently task-dependent. While all nine evaluated datasets eventually converge to a low-VIG saturated state, their evolutionary trajectories vary substantially. For instance, fine-grained visual recognition tasks (e.g., VizWiz, GQA) exhibit different deep-layer attention fluctuations compared to hallucination-sensitive tasks (e.g., POPE, which drops sharply at layer 19). This compelling empirical evidence directly invalidates the use of a monolithic, one-size-fits-all pruning mask (e.g., ShortV, VTW). Instead, it firmly justifies our few-shot calibration strategy, which elegantly tailors the optimal skipping subset to the unique semantic reasoning pathways required by each specific downstream task.

\subsection{V-Skip: Block-wise Structured Sparsity}
\label{subsec:vskip}

Based on the saturation analysis, we propose \textbf{V-Skip}, a training-free acceleration mechanism that imposes structured sparsity on the attention mechanism in deep layers.
To ensure robustness across different architectures and tasks, we adopt a data-driven selection strategy rather than fixed heuristics.

\noindent\textbf{Identification of Saturated Layers via  Top-N.} To maximize efficiency under a computational budget, we identify the top-N most saturated layers to bypass.
Formally, let $\mathcal{L}_{all} = \{0, 1, \dots, L-1\}$ be the set of all decoder layers. We seek a subset $\mathcal{L}_{skip}^{(k)} \subset \mathcal{L}_{all}$ with cardinality $|\mathcal{L}_{skip}^{(k)}| = \text{N}$ such that the total information loss is minimized:
\begin{equation}
\mathcal{L}_{skip}^{(k)} = \mathop{\arg\min}_{\mathcal{S}\subset\mathcal{L}_{all},|\mathcal{S}|=\text{N}} \sum_{l\in\mathcal{S}}\mathcal{G}_k(l)
    \label{eq:topk}
\end{equation}
This optimization effectively selects the N layers with the lowest $\mathcal{G}_k(l)$ scores. By adjusting N, V-Skip offers a flexible trade-off between speed and precision.

\noindent\textbf{Decoupled Attention Masking.}
Let $\mathcal{L}_{skip}^{(k)}$ denote the set of saturated layers (e.g., layers 12 to 31). For any layer $l \in \mathcal{L}_{skip}^{(k)}$, we modify the standard causal attention mask $\mathbf{M} \in \{0, -\infty\}^{N \times N}$ to a block-sparse mask $\mathbf{M}_{skip}$.
Specifically, the standard attention score calculation is:
\begin{equation}
    \mathbf{S} = \text{Softmax}\left(\frac{\mathbf{Q}\mathbf{K}^T}{\sqrt{d}} + \mathbf{M}\right).
\end{equation}
In V-Skip, we alter the visibility of visual tokens. For query tokens $i \in \mathcal{V}$ (visual tokens), we enforce a bypass mechanism, effectively eliminating the expensive query-key product:
\begin{equation}
    \mathbf{H}'^l_{\mathcal{V}} \longleftarrow \text{LN}(\mathbf{H}^l_{\mathcal{V}}) + \mathbf{H}^l_{\mathcal{V}}. \quad (\text{Visual Bypass})
\end{equation}
For query tokens $j \in \mathcal{T}$ (textual tokens), the attention remains fully accessible to both visual and textual contexts to preserve reasoning capabilities:
\begin{equation}
    \mathbf{h}'^l_j = \sum_{k \in \mathcal{V} \cup \mathcal{T}} \text{Attn}(q_j, k_k) \cdot v_k. \quad (\text{Textual Reasoning})
\end{equation}
This formulation is equivalent to applying a heterogeneous attention mask where visual queries cannot attend to other visual tokens, but textual queries retain full visibility.

\noindent\textbf{FFN Preservation for Alignment.}
Crucially, distinct from layer-skipping approaches (e.g., ShortV), V-Skip retains the FFN operation for \textit{all} tokens:
\begin{equation}
    \mathbf{H}^{l+1} = \text{FFN}(\text{LN}(\mathbf{H}'^l)) + \mathbf{H}'^l.
\end{equation}
This design choice is theoretically grounded in our observation that while spatial mixing saturates, the point-wise non-linear projection $f: \mathbb{R}^d \to \mathbb{R}^d$ performed by the FFN is essential for maintaining the semantic trajectory of visual tokens within the LLM's manifold.

\noindent\textbf{Complexity Analysis.}
Standard MHSA requires $\mathcal{O}((N_v + N_t)^2)$ operations. V-Skip reduces the complexity in skipped layers to $\mathcal{O}(N_t^2 + N_t \cdot N_v)$. Since $N_v \gg N_t$, the dominant quadratic term $N_v^2$ is eliminated. The theoretical speedup ratio for the attention module approaches $ N_v /N_t + 1$, offering significant latency reduction for high-resolution inputs.

\section{Experiments}
\label{experiments}

\subsection{Experimental Setup}
We evaluate our method on representative MLLM architectures, specifically LLaVA-1.5 (7B/13B) \cite{llava}, LLaVA-NeXT (7B/13B) \cite{llavanext} and Qwen2.5-VL-7B \cite{qwenvl25}. LLaVA-1.5 processes $336\times336$ images resulting in 576 visual tokens, whereas LLaVA-NeXT employs ``AnyRes'' tiling to scale the visual sequence up to 2,880 tokens per image. This extended sequence length explicitly exposes the quadratic bottleneck of visual self-attention. All evaluations are conducted using the standardized lmms-eval framework, strictly adhering to its official protocols. Our experiments were executed on an NVIDIA RTX 3090 GPU.

To validate our structural optimization paradigm, we compare V-Skip against representative training-free methods: VTW\cite{vtw} and ShortV\cite{shortv}. These methods are selected as they also operate on depth-wise structural modifications. However, they inherently differ from our routing mechanism: VTW destructively discards all visual tokens after a predefined layer $K$, while ShortV discretely skips entire LLM layers (losing critical FFN transformations). For fair comparison, we adopt the default configurations reported in their respective works: $K{=}16$ (7B) and $K{=}20$ (13B) for VTW; $\text{N}{=}19$ (7B) and $\text{N}{=}24$ (13B) for ShortV.

\noindent\textbf{Benchmarks.} To evaluate the general and fine-grained capabilities of V-Skip, we conduct experiments on a comprehensive suite of nine multimodal benchmarks. These include general evaluation (MME~\cite{mme}, MMBench~\cite{mmb}), expert-level knowledge and reasoning (MMMU~\cite{mmmu}, GQA~\cite{gqa}, ScienceQA~\cite{sqa}), object hallucination sensitivity (POPE~\cite{pope}), and specialized tasks such as structured diagram understanding (AI2D~\cite{AI2D}), in-the-wild VQA (VizWiz~\cite{VizWiz}), and fine-grained text recognition (OCRBench~\cite{OCRBench}).

\subsection{Main Results}
\label{sec:main_results}

The performance comparison of V-Skip against representative training-free methods is summarized in~\cref{tab:main_comparison}. Overall, V-Skip consistently achieves superior accuracy retention across all evaluated architectures, spanning the LLaVA series to the state-of-the-art Qwen2.5-VL-7B. By effectively bypassing redundant visual self-attention while preserving FFN-based semantic evolution, V-Skip maintains a robust average performance retention of 98.42\% to 100.31\% on the LLaVA series, and achieves a highly competitive 94.16\% on the Qwen architecture.

\noindent\textbf{Superiority over Destructive Pruning.} As observed on LLaVA-1.5-7B, V-Skip achieves a near-lossless average retention of 100.31\%, even slightly surpassing the vanilla model on tasks like MMBench and OCRBench. In contrast, token pruning methods such as VTW suffer from catastrophic degradation on fine-grained tasks (e.g., dropping from 31.3 to 5.1 on OCRBench). This highlights that preserving full visual context is essential for complex multimodal reasoning. V-Skip satisfies this through a non-destructive architectural design, eliminating redundancy while maintaining the integrity of the visual tokens.

\noindent\textbf{Generalizability across Scales and MLLM Paradigms.} V-Skip consistently maintains near-lossless accuracy on 13B models, achieving 99.77\% and 99.75\% retention on LLaVA-1.5-13B and LLaVA-NeXT-13B, respectively. It remains robust on the highly optimized Qwen2.5-VL-7B with 94.16\% retention, significantly outperforming the layer-skipping approach ShortV (87.11\%). These results prove that skipping visual attention saturation blocks is a robust strategy compared to token withdrawal or layer freezing.

\begin{table}[t]
    \centering
    \setlength{\tabcolsep}{3.5pt}
    \footnotesize
    \caption{Performance comparison of various methods across different benchmarks. Best results in \textcolor{red}{\textbf{red}}, and second best results in \textcolor{MidnightBlue}{\textbf{blue}}.}
    \label{tab:main_comparison}
    \resizebox{\linewidth}{!}{
    \begin{tabular}{l | c c c c c c c c c | c}
        \hline
        \textbf{\;Methods} & \textbf{MME} & \textbf{MMMU} & \textbf{MMB} & \textbf{GQA} & \textbf{POPE} & \textbf{SQA} & \textbf{AI2D} & \textbf{VizWiz} & \textbf{OCRB} & \makecell{\textbf{Avg.}}\\
        \hline
        
       \rowcolor{gray!15} LLaVA-1.5-7B & 1866.1 & 36.7 & 64.1 & 61.9 & 85.1 & 69.6 & 55.2 & 54.3 & 31.3 & 100.00\%\\
       VTW(K=16) & \textcolor{red}{\textbf{1857.6}} & \textcolor{MidnightBlue}{\textbf{36.4}} & 64.0 & 55.1 & \textcolor{MidnightBlue}{\textbf{85.3}} & \textcolor{red}{\textbf{69.6}} & \textcolor{red}{\textbf{55.4}} & \textcolor{MidnightBlue}{\textbf{51.0}} & 5.1 & 88.71\%\\
       ShortV(N=19) & 1842.2 & 36.2 & \textcolor{MidnightBlue}{\textbf{64.8}} & \textcolor{MidnightBlue}{\textbf{60.8}} & 84.4 & 68.3 & 54.2 & 50.6 & \textcolor{MidnightBlue}{\textbf{29.5}} & \textcolor{MidnightBlue}{\textbf{97.73\%}}\\
       V-Skip(N=19) & \textcolor{MidnightBlue}{\textbf{1853.3}} & \textcolor{red}{\textbf{36.9}} & \textcolor{red}{\textbf{65.0}} & \textcolor{red}{\textbf{61.8}} & \textcolor{red}{\textbf{85.5}} & \textcolor{MidnightBlue}{\textbf{69.3}} & \textcolor{red}{\textbf{55.4}} & \textcolor{red}{\textbf{54.5}} & \textcolor{red}{\textbf{31.6}} & \textcolor{red}{\textbf{100.31\%}}\\
        \hline

       \rowcolor{gray!15} LLaVA-1.5-13B & 1824.2 & 35.7 & 68.7 & 63.3 & 85.6 & 72.7 & 59.3 & 56.6 & 33.7 & 100.00\%\\
       VTW(K=20) & \textcolor{MidnightBlue}{\textbf{1828.8}} & 34.9 & \textcolor{red}{\textbf{68.8}} & 60.6 & \textcolor{red}{\textbf{87.1}} & \textcolor{red}{\textbf{72.9}} & \textcolor{red}{\textbf{59.4}} & 55.1 & 24.2 & 96.14\%\\
       ShortV(N=24) & 1828.7 & \textcolor{red}{\textbf{35.7}} & \textcolor{MidnightBlue}{\textbf{68.7}} & \textcolor{red}{\textbf{63.3}} & \textcolor{MidnightBlue}{\textbf{85.6}} & \textcolor{MidnightBlue}{\textbf{72.7}} & \textcolor{MidnightBlue}{\textbf{59.3}} & \textcolor{MidnightBlue}{\textbf{56.6}} & \textcolor{MidnightBlue}{\textbf{32.1}} & \textcolor{MidnightBlue}{\textbf{99.50\%}}\\
       V-Skip(N=24) & \textcolor{red}{\textbf{1831.9}} & \textcolor{MidnightBlue}{\textbf{35.4}} & \textcolor{MidnightBlue}{\textbf{68.7}} & \textcolor{MidnightBlue}{\textbf{62.8}} & 85.5 & \textcolor{red}{\textbf{72.9}} & 59.2 & \textcolor{red}{\textbf{57.3}} & \textcolor{red}{\textbf{33.0}} & \textcolor{red}{\textbf{99.77\%}}\\
        \hline

       \rowcolor{gray!15} LLaVA-NeXT-7B & 1846.3 & 36.7 & 67.2 & 64.3 & 86.4 & 70.2 & 65.3 & 60.6 & 52.3 & 100.00\%\\
       VTW(K=16) & \textcolor{red}{\textbf{1853.9}} & \textcolor{red}{\textbf{36.8}} & \textcolor{MidnightBlue}{\textbf{67.1}} & \textcolor{red}{\textbf{63.7}} & \textcolor{red}{\textbf{86.4}} & 66.5 & \textcolor{red}{\textbf{65.6}} & \textcolor{red}{\textbf{59.6}} & 6.5 & 89.51\%\\
       ShortV(N=19) & \textcolor{MidnightBlue}{\textbf{1844.1}} & 35.8 & \textcolor{red}{\textbf{67.2}} & \textcolor{MidnightBlue}{\textbf{63.5}} & 86.2 & \textcolor{MidnightBlue}{\textbf{69.1}} & 64.4 & 57.7 & \textcolor{MidnightBlue}{\textbf{45.6}} & \textcolor{MidnightBlue}{\textbf{97.27\%}}\\
       V-Skip(N=19) & 1778.6 & \textcolor{MidnightBlue}{\textbf{36.6}} & 66.8 & 63.4 & \textcolor{MidnightBlue}{\textbf{86.3}} & \textcolor{red}{\textbf{70.0}} & \textcolor{MidnightBlue}{\textbf{65.2}} & \textcolor{MidnightBlue}{\textbf{59.3}} & \textcolor{red}{\textbf{49.4}} & \textcolor{red}{\textbf{98.42\%}}\\
        \hline

       \rowcolor{gray!15} LLaVA-NeXT-13B & 1892.0 & 34.9 & 69.2 & 65.4 & 86.4 & 73.6 & 70.3 & 63.6 & 55.1 & 100.00\%\\
       VTW(K=20) & 1878.1 & \textcolor{red}{\textbf{36.1}} & \textcolor{MidnightBlue}{\textbf{69.2}} & 61.4 & \textcolor{red}{\textbf{87.4}} & \textcolor{MidnightBlue}{\textbf{73.4}} & \textcolor{MidnightBlue}{\textbf{70.2}} & 58.5 & 35.5 & 94.86\%\\
       ShortV(N=24) & \textcolor{red}{\textbf{1899.6}} & \textcolor{red}{\textbf{36.1}} & 69.1 & \textcolor{MidnightBlue}{\textbf{63.6}} & 86.2 & 73.1 & 69.5 & \textcolor{MidnightBlue}{\textbf{59.2}} & \textcolor{MidnightBlue}{\textbf{49.6}} & \textcolor{MidnightBlue}{\textbf{98.00\%}}\\
       V-Skip(N=24) & \textcolor{MidnightBlue}{\textbf{1885.4}} & \textcolor{MidnightBlue}{\textbf{35.3}} & \textcolor{red}{\textbf{69.3}} & \textcolor{red}{\textbf{64.9}} & \textcolor{MidnightBlue}{\textbf{86.5}} & \textcolor{red}{\textbf{73.9}} & \textcolor{red}{\textbf{70.5}} & \textcolor{red}{\textbf{62.9}} & \textcolor{red}{\textbf{53.9}} & \textcolor{red}{\textbf{99.75\%}}\\
        \hline
        
        \rowcolor{gray!15} Qwen2.5-VL-7B & 2344.1 & 50.4 & 83.6 & 60.5 & 86.4 & 76.4 & 82.7 & 70.2 & 84.4 & 100.00\% \\
        VTW(K=7) & \textcolor{red}{\textbf{2319.9}} & \textcolor{red}{\textbf{46.8}} & \textcolor{MidnightBlue}{\textbf{72.8}} & \textcolor{red}{\textbf{56.5}} & \textcolor{red}{\textbf{86.2}} & 75.3 & \textcolor{MidnightBlue}{\textbf{76.4}} & \textcolor{red}{\textbf{66.4}} & 74.6 & \textcolor{MidnightBlue}{\textbf{94.00\%}} \\
        ShortV(N=7) & 2092.1 & 42.8 & \textcolor{MidnightBlue}{\textbf{72.8}} & \textcolor{MidnightBlue}{\textbf{54.6}} & 84.0 & \textcolor{MidnightBlue}{\textbf{78.3}} & 55.2 & 50.0 & \textcolor{MidnightBlue}{\textbf{80.0}} & 87.11\% \\
        V-Skip(N=7) & \textcolor{MidnightBlue}{\textbf{2228.8}} & \textcolor{MidnightBlue}{\textbf{43.3}} & \textcolor{red}{\textbf{74.7}} & 51.3 & \textcolor{red}{\textbf{86.2}} & \textcolor{red}{\textbf{81.7}} & \textcolor{red}{\textbf{76.5}} & \textcolor{red}{\textbf{66.4}} & \textcolor{red}{\textbf{83.1}} & \textcolor{red}{\textbf{94.16\%}} \\
        \hline
    \end{tabular}}
\end{table}

\subsection{V-Skip as a Universal Plug-and-Play Booster}
To verify that V-Skip is complementary to existing acceleration techniques, we integrate it with FastV, a representative visual token pruning method. We set the pruning layer index $K$ to 3 and the keep token pruning ratio to 0.5. For our module, we configure the model with a sparsity budget of N=19. As presented in ~\cref{tab:orth_fastv_main}, while FastV efficiently reduces token count, it inevitably introduces subtle performance trade-offs relative to the dense baseline. Remarkably, integrating V-Skip as a booster not only further compresses the computational load but also actively restores and even enhances the reasoning capability across all nine benchmarks. Specifically, compared to the FastV, the combined FastV+V-Skip improves MME by +10.0, MMMU by +0.4, POPE by +0.2, and OCRBench by +0.6. From an efficiency perspective, V-Skip further reduces the computational requirement from 61.66\% to 56.73\% TFLOPs of the vanilla model. These results demonstrate that V-Skip serves as a powerful plug-and-play component. While token pruning reduces the input quantity, V-Skip optimizes the visual interactions within the attention blocks. This synergy allows for multi-dimensional sparsity while simultaneously safeguarding the model's reasoning capability.

\begin{table}[t]
    \centering
    \setlength{\tabcolsep}{3.5pt}
    \footnotesize
    \caption{Integrating V-Skip with FastV to enhance both reasoning capability and efficiency on LLaVA-1.5-7B.}
    \label{tab:orth_fastv_main}
    \resizebox{\linewidth}{!}{
        \begin{tabular}{l | c c c c c c c c c | c}
            \hline
            \textbf{Method} & \textbf{MME} & \textbf{MMMU} & \textbf{MMB} & \textbf{GQA} & \textbf{POPE} & \textbf{SQA} & \textbf{AI2D} & \textbf{VizWiz} & \textbf{OCRB} & \makecell{\textbf{TFLOPs}} \\
            \hline
            \rowcolor{gray!15} 
            LLaVA-1.5-7B       & 1866.1       & 36.7         & 64.1         & 61.9         & 85.1          & 69.6         & 55.2          & 54.3           & 31.3         & 100.00\% \\
            FastV         & 1863.8       & 35.8         & \textbf{63.8} & \textbf{60.4} & 83.9 & 68.7 & 55.1 & 54.5 & 30.6 & 61.66\% \\
            FastV+V-Skip    & \textbf{1873.8} & \textbf{36.2} & 63.4 & 60.1 & \textbf{84.1} & \textbf{69.0} & \textbf{55.2} & \textbf{54.7} & \textbf{31.2} & \textbf{56.73\%} \\
            \hline
        \end{tabular}
    }
\end{table}

\newpage
\subsection{Analyzing the Efficiency-Effectiveness Balance}
\begin{wraptable}{r}{0.49\textwidth}
    \vspace{-1.05\baselineskip}
    \centering
    \setlength{\tabcolsep}{2.5pt}
    \scriptsize
    \resizebox{\linewidth}{!}{
     \begin{tabular}{l | c c l} 
            \toprule
            \textbf{Method} & \textbf{TFLOPs} & \textbf{Latency} & \textbf{Avg.} ($\Delta$) \\
            \midrule
            \rowcolor{gray!15} 
            LLaVA-1.5-7B  & 8.54  & 171.71 & 100.00\% \\
            V-Skip (N=19) & 7.69  & 158.40 & \textbf{100.31\%} (\textcolor{red}{\textbf{$\uparrow$ 0.31}})\\
            VTW (K=16)    & 4.94  & 107.57 & 88.71\% (\textcolor{blue}{$\downarrow$ 11.29})\\
            ShortV (N=19) & 4.72  & 108.40 & 97.73\% (\textcolor{blue}{$\downarrow$ 2.27})\\
            \midrule
            \rowcolor{gray!15} 
            LLaVA-NeXT-7B & 30.74 & 527.34 & 100.00\% \\
            V-Skip (N=19) & 26.57 & 486.50 & \textbf{98.42\%} (\textcolor{red}{\textbf{$\downarrow$ 1.58}})\\
            VTW (K=16)    & 16.91 & 366.21 & 89.51\% (\textcolor{blue}{$\downarrow$ 10.49})\\
            ShortV (N=19) & 15.73 & 332.22 & 97.27\% (\textcolor{blue}{$\downarrow$ 2.73})\\
            \bottomrule
        \end{tabular}}
    \caption{Performance and efficiency comparison across MLLM architectures.}
    \label{tab:performance_comparison}
\end{wraptable}
We analyze the trade-off between theoretical complexity (TFLOPs), inference latency, and accuracy retention. As summarized in~\cref{tab:performance_comparison}, V-Skip establishes a compelling efficiency-effectiveness trade-off by effectively mitigating the quadratic bottleneck of long visual sequences while preserving the model's ability. On LLaVA-1.5-7B, V-Skip reduces TFLOPs from 8.54 to 7.69 and achieves a 13.31 ms reduction in latency. Crucially, unlike compared methods, V-Skip uniquely attains a +0.31\% average performance gain, proving that eliminating redundant attention can even serve as a beneficial regularizer. This trend holds on the higher-resolution LLaVA-NeXT-7B, where V-Skip maintains the highest retention (98.42\%) among all methods, significantly outperforming the destructive pruning of VTW (89.51\%).

While VTW and ShortV offer more aggressive reductions in TFLOPs, they suffer from catastrophic degradation (up to -11.29\%) due to their disruption of the model's architectural integrity. We argue that preserving FFN transformations is an essential investment; while bypassing the $\mathcal{O}(N^2)$ visual self-attention captures the bulk of computational redundancy, the FFNs remain vital for maintaining the semantic trajectory of visual tokens. This strategic decoupling allows V-Skip to provide meaningful acceleration while ensuring the model remains robust for complex multimodal tasks.

\subsection{Ablation Studies}

\subsubsection{Decoupling Spatial Interaction from Semantic Evolution.}
To validate our functional decoupling strategy, we investigate whether deep-layer redundancy in MLLMs primarily resides in spatial interaction or semantic projection. We compare V-Skip against an FFN-Skip strategy, where visual tokens bypass FFN transformations but continue to participate in self-attention, effectively halting their non-linear evolution while maintaining quadratic spatial mixing.

As presented in~\cref{tab:layer_skipping}, V-Skip demonstrates superior robustness with an average performance retention of 100.31\% compared to 99.16\% for FFN-Skip. While FFN-Skip shows marginal gains on MME and ScienceQA, it suffers noticeable degradations on tasks requiring fine-grained visual grounding, such as OCRBench and the hallucination-sensitive POPE benchmark. These results corroborate that bypassing the FFN block prevents visual tokens from undergoing the necessary layer-wise semantic evolution required for high-level reasoning. Consequently, V-Skip successfully eliminates spatial redundancy through attention skipping while preserving the critical semantic alignment and trajectory essential for complex multimodal understanding.

\begin{table*}[!b]
    \centering
    \caption{Ablation on functional decoupling. Comparing V-Skip against FFN-Skip to demonstrate the necessity of preserving FFN transformations for semantic evolution.}
    \label{tab:layer_skipping}
    \resizebox{\textwidth}{!}{
    \begin{tabular}{@{} c c c ccccccccc @{\quad} c @{}}
        \toprule
        \multirow{2}{*}{\textbf{Method}} & \multicolumn{2}{c}{\textbf{DecoderLayer}} & \multirow{2}{*}{\textbf{MME}} & \multirow{2}{*}{\textbf{MMMU}} & \multirow{2}{*}{\textbf{MMB}} & \multirow{2}{*}{\textbf{GQA}} & \multirow{2}{*}{\textbf{POPE}} & \multirow{2}{*}{\textbf{SQA}} & \multirow{2}{*}{\textbf{AI2D}} & \multirow{2}{*}{\textbf{VizWiz}} & \multirow{2}{*}{\textbf{OCRB}} & \multirow{2}{*}{\textbf{Avg.}} \\
        \cmidrule(lr){2-3} 
        & \textbf{MHSA} & \textbf{FFN} & & & & & & & & & & \\ 
        \midrule
        \textcolor{gray}{V-Skip}   & \checkmark &            & 1853.3 & \textbf{36.9} & \textbf{65.0} & \textbf{61.8} & \textbf{85.5} & 69.3 & \textbf{55.4} & \textbf{54.5} & \textbf{31.6} & \textbf{100.31\%} \\
        \textcolor{gray}{FFN-Skip} &            & \checkmark & \textbf{1873.9} & 36.0 & 64.4 & 61.0 & 85.1 & \textbf{69.7} & 55.2 & 53.2 & 30.3 & 99.16\%  \\
        \textcolor{gray}{ShortV}   & \checkmark & \checkmark & 1842.2 & 36.2 & 64.8 & 60.8 & 84.4 & 68.3 & 54.2 & 50.6 & 29.5 & 97.73\% \\ 
        \bottomrule
    \end{tabular}
    }
\end{table*}

\begin{table}[!t]
    \centering
    \setlength{\tabcolsep}{3.5pt}
    \footnotesize
    \caption{Sensitivity to sparsity budget N. Performance across varying budgets on LLaVA-1.5-7B, highlighting the near-lossless regime and the optimal balance at N=19.}
    \label{tab:V-Skip_increasing}
    \resizebox{\linewidth}{!}{
    \begin{tabular}{c | c c c c c c c c c | c}
        \hline
        \textbf{\;Budget(N)} & \textbf{MME} & \textbf{MMMU} & \textbf{MMB} & \textbf{GQA} & \textbf{POPE} & \textbf{SQA} & \textbf{AI2D} & \textbf{VizWiz} & \textbf{OCR} & \makecell{\textbf{Avg.}}\\
        \hline
        \rowcolor{gray!15} 0 & 1866.1 & 36.7 & 64.1 & 61.9 & 85.1 & 69.6 & 55.2 & 54.3 & 31.3 & 100.00\% \\
        4 & 1858.4 & 36.4 & 64.1 & 61.9 & 85.2 & 69.6 & 55.2 & 54.0 & 31.4 & 99.85\% \\
        8 & 1849.8 & 36.7 & 64.2 & 61.8 & 85.2 & 69.3 & 55.2 & 54.1 & 31.6 & 99.93\% \\
        16 & 1844.7 & 36.7 & 64.3 & 61.8 & 85.3 & 69.3 & 55.4 & 54.1 & 31.6 & 99.97\% \\
        19 & 1853.3 & 36.9 & 65.0 & 61.8 & 85.5 & 69.3 & 55.4 & 54.5 & 31.6 & 100.31\% \\
        21 & 1834.7 & 36.7 & 64.6 & 61.6 & 85.3 & 69.4 & 55.1 & 53.9 & 29.9 & 99.24\% \\
        24 & 1807.4 & 36.2 & 64.7 & 61.3 & 85.1 & 69.1 & 55.0 & 54.6 & 29.4 & 98.76\% \\
        32 & 1560.1 & 35.2 & 60.7 & 54.5 & 78.7 & 68.8 & 52.2 & 54.8 & 22.1 & 91.08\% \\
        \hline
    \end{tabular}}
\end{table}

\subsubsection{Impact of Attention Sparsity Budget N.}
We investigate the sensitivity of model performance to the attention sparsity budget N, which defines the cardinality of the skipped layer subset $|\mathcal{L}_{skip}^{(k)}|=\text{N}$ . By progressively increasing N from 0 to 32, we evaluate the trade-off between computational efficiency and reasoning accuracy on LLaVA-1.5-7B across all benchmarks. Following our evaluation protocol, we report the average retention percentage relative to the vanilla dense baseline ($\text{N}=0$).
As illustrated in~\cref{tab:V-Skip_increasing}, V-Skip demonstrates remarkable robustness within a broad near-lossless regime. Specifically, setting the budget N between 4 and 19 yields an average retention of at least 99.85\%. Notably, the configuration with N=19 even slightly surpasses the baseline performance, achieving a peak retention of 100.31\%. This validates our core hypothesis that a substantial portion of visual self-attention in deeper layers is indeed computationally redundant and can be safely bypassed without degrading model capabilities.
However, performance begins to decline when the budget exceeds the identified saturation threshold. Accuracy retention drops to 99.24\% at N=21 and 98.76\% at N=24. Aggressively skipping the attention modules in all 32 layers (N=L) results in a significant degradation to 91.08\%, particularly on fine-grained tasks such as MME, GQA, and OCRBench. This suggests that excessive skipping eventually removes critical visual features required for localized recognition and complex reasoning. Based on this analysis, we select N=19 (representing approximately 60\% of the total layers) as our default configuration to achieve an optimal balance between architectural sparsity and preservation of precision.

\begin{figure}[t]
    \centering
    \begin{minipage}{0.39\linewidth}
        \centering
        \includegraphics[width=\linewidth]{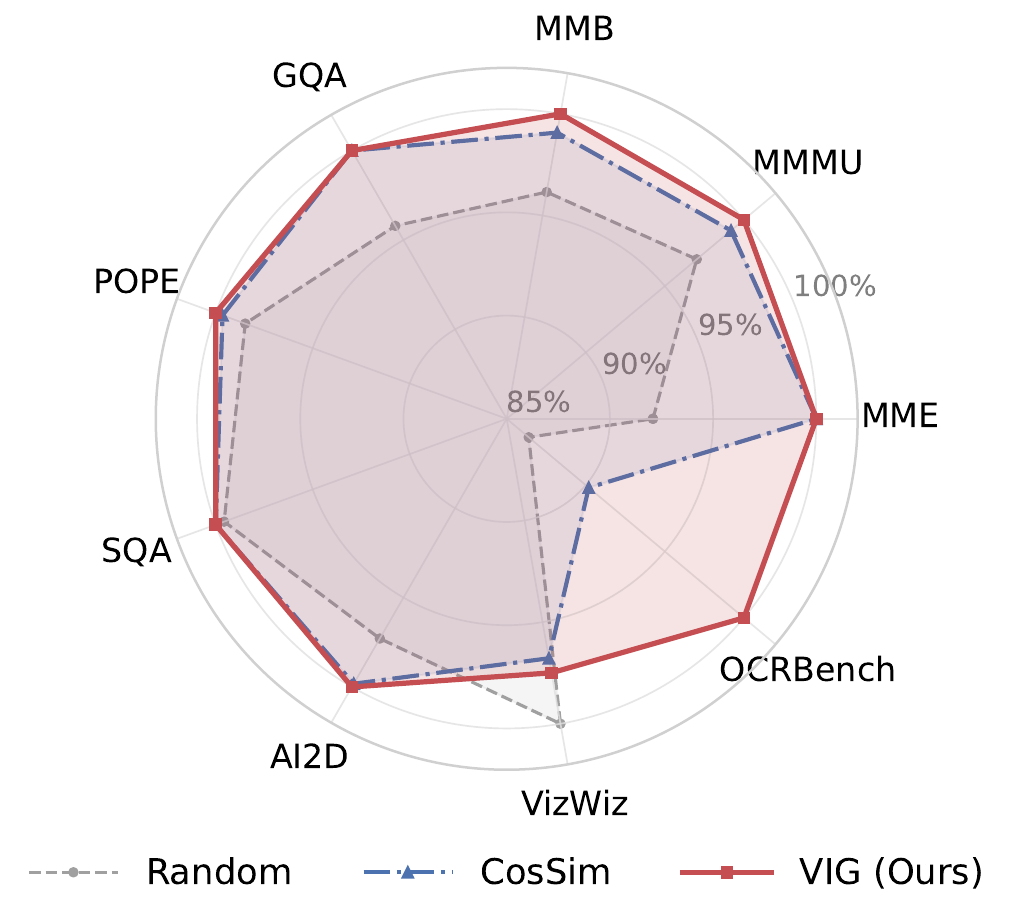}  
    \caption{Performance retention of different identification strategies.}
    \label{fig:strategy_comparison}
    \end{minipage}
    \hfill 
    \begin{minipage}{0.6\linewidth}
        \centering
        \includegraphics[width=\linewidth]{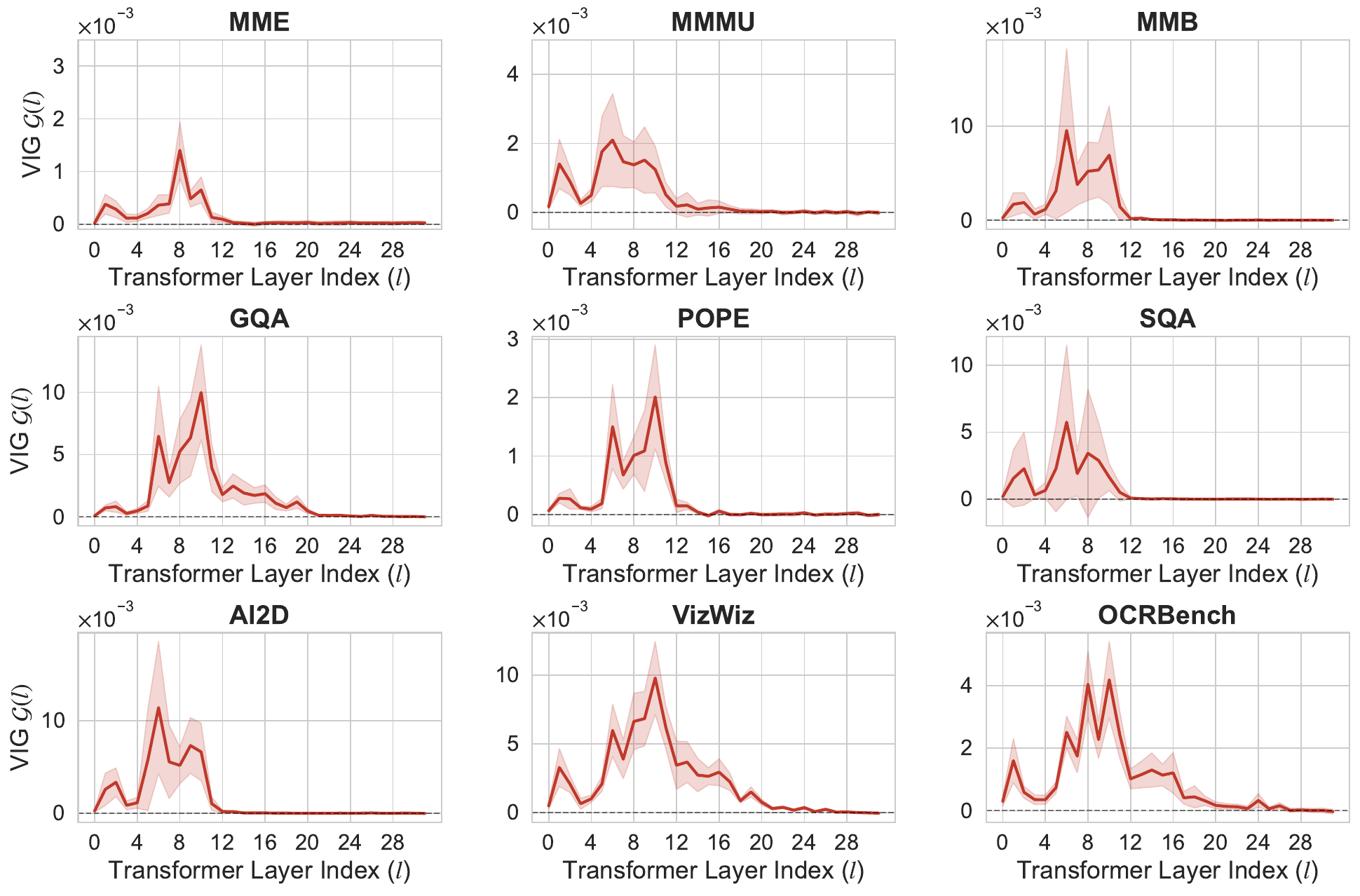} 
  \caption{Robustness analysis of the few-shot calibration mechanism. VIG $\mathcal{G}(l)$ exhibits consistent task-level trends across different data sampling seeds.}
  \label{fig:calibrationSet}
    \end{minipage}
\end{figure}

\subsubsection{Impact of Identification Strategy.} 
\label{subsec:identification_ablation_strategy}

To validate the VIG metric, we compare it to Random selection and Cosine Similarity (CosSim) strategies under an identical sparsity budget of N=19. The CosSim baseline is calculated directly from output logits. As shown in \cref{fig:strategy_comparison}, the red line indicates that VIG yields the most robust performance envelope and maintains consistent competitiveness across diverse benchmarks. While CosSim outperforms Random skipping, its performance fluctuates significantly and falls short on structure-sensitive tasks like OCRBench. Consequently, VIG delivers a superior trade-off across multi-task evaluations. By measuring redundancy via visual information divergence instead of the geometric angle of logits, VIG highly accurately identifies semantically inert visual attention saturation blocks. This approach ensures robust results in both general and fine-grained contexts.

\subsubsection{Robustness of Few-shot Calibration.}
To rigorously validate the stability of our task-aware calibration, we investigate its sensitivity to the random sampling of few-shot examples. \cref{fig:calibrationSet} illustrates the Visual Information Gap ($\mathcal{G}(l)$) across all transformer layers for nine diverse benchmarks. The solid red lines represent the mean $\mathcal{G}(l)$ averaged over multiple random seeds, while the shaded regions denote the variance. The results show that while early layers have sample-dependent variance, deep layers maintain near-zero variance across all random seeds and tasks. This firmly demonstrates that Visual Attention Saturation is an inherent, dataset-level invariant rather than an artifact of specific few-shot samples. Consequently, our highly efficient 20-shot calibration mechanism is statistically robust and insensitive to the randomness of the calibration subset, ensuring stable and reliable skip routing decisions for varying downstream tasks.

\begin{figure}[t] 
    \centering
    
    \begin{minipage}[b]{0.42\textwidth}
        \centering
        \includegraphics[width=\linewidth]{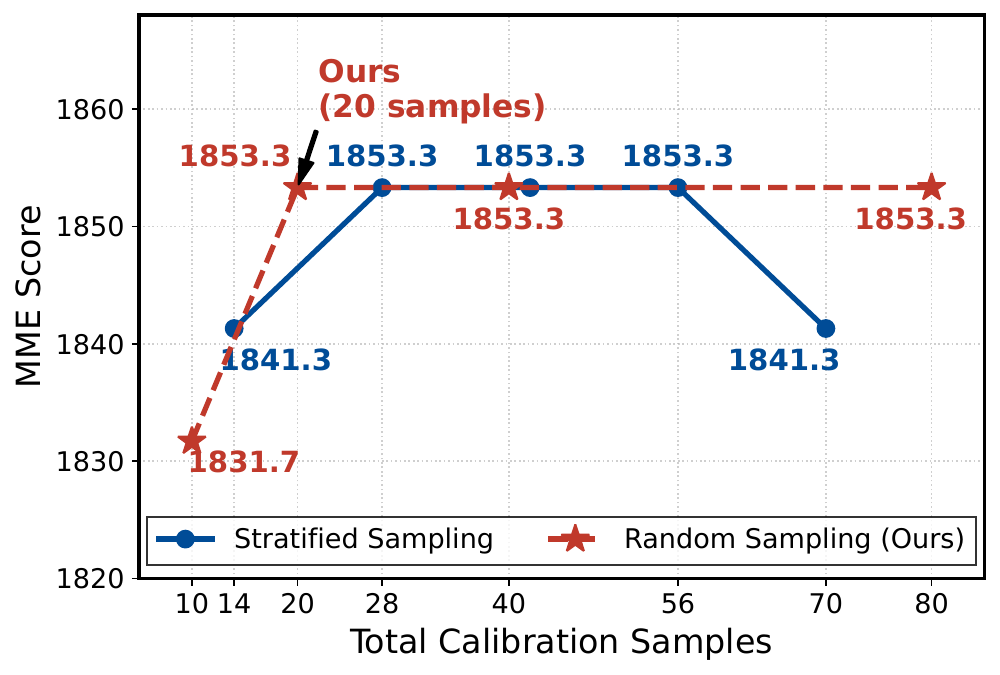}
      \caption{Sensitivity to calibration sampling strategies.}
      \label{fig:samplingAb}
    \end{minipage}
    \hfill
    \begin{minipage}[b]{0.55\textwidth}
        \centering
    \footnotesize
        \resizebox{0.75\linewidth}{!}{
\begin{tabular}{ccc}
    \toprule
    \textbf{Target} & \textbf{MMBench} & \textbf{OCRBench} \\
    \midrule
    \textbf{Domain}   & General (Similar) & Text (Distant) \\
    \textbf{Native}   & \textbf{65.00}     & \textbf{31.60}  \\
    \textbf{Transfer} & 64.43             & 30.70          \\
    \rowcolor{gray!15} 
    \textbf{$\Delta$} & \textcolor{blue}{$\downarrow$ 0.57} & \textcolor{blue}{\textbf{$\downarrow$ 0.90}} \\
    \bottomrule
\end{tabular}
        }
    \caption{Cross-dataset transferability. ``Native'' denotes using the 20-shot strategy calibrated on the target benchmark, while ``Transfer'' applies the MME-calibrated strategy. The performance drop on the text-centric OCRBench highlights the necessity of task-specific calibration.}
    \label{tab:cross_dataset_transfer}
    \end{minipage}
    
\end{figure}

\subsubsection{Sensitivity to Sub-task Distribution.}
Building upon the seed-level stability, we further investigate whether the calibration process requires balanced data to cover complex, multi-task scenarios. Using the MME benchmark as a representative case study, which comprises 14 distinct sub-tasks, we compare our default global random sampling against a task-stratified sampling strategy. As illustrated in \cref{fig:samplingAb}, the stratified approach plateaus at a peak score of 1853.3 when utilizing 28 to 56 total samples (corresponding to $m \in \{2,3,4\}$ instances per sub-task). Remarkably, our global random sampling achieves this identical performance 1853.3 with a budget of only 20 samples. This confirms that for a general-purpose benchmark like MME, its diverse sub-tasks collectively share a stable macroscopic saturation profile. A minimal random subset easily captures this global average without requiring meticulous sub-task balancing.

However, this internal sub-task robustness does not imply universal transferability across fundamentally different domains. To demonstrate the strict necessity of macro-level, task-specific calibration, we evaluate cross-dataset transferability as shown in \cref{tab:cross_dataset_transfer}. Applying the routing strategy calibrated on MME to the structurally similar MMBench yields a highly competitive score of 64.43 (vs. 65.0 native). In contrast, transferring it to OCRBench, a specialized, extreme domain featuring dense text and complex structural parsing, results in a noticeable performance drop (30.7 vs. 31.6 native). 
While the saturation depth is highly robust to micro-level sub-task variations within a general domain (making our 20-shot random sampling extremely efficient), it shifts drastically when facing a specialized domain gap. Consequently, an independent, ultra-lightweight calibration for each distinct downstream task is not just practical, but essential.

\subsection{Qualitative Analysis} 

\begin{figure}[tb]
  \centering
  \includegraphics[width=\linewidth]{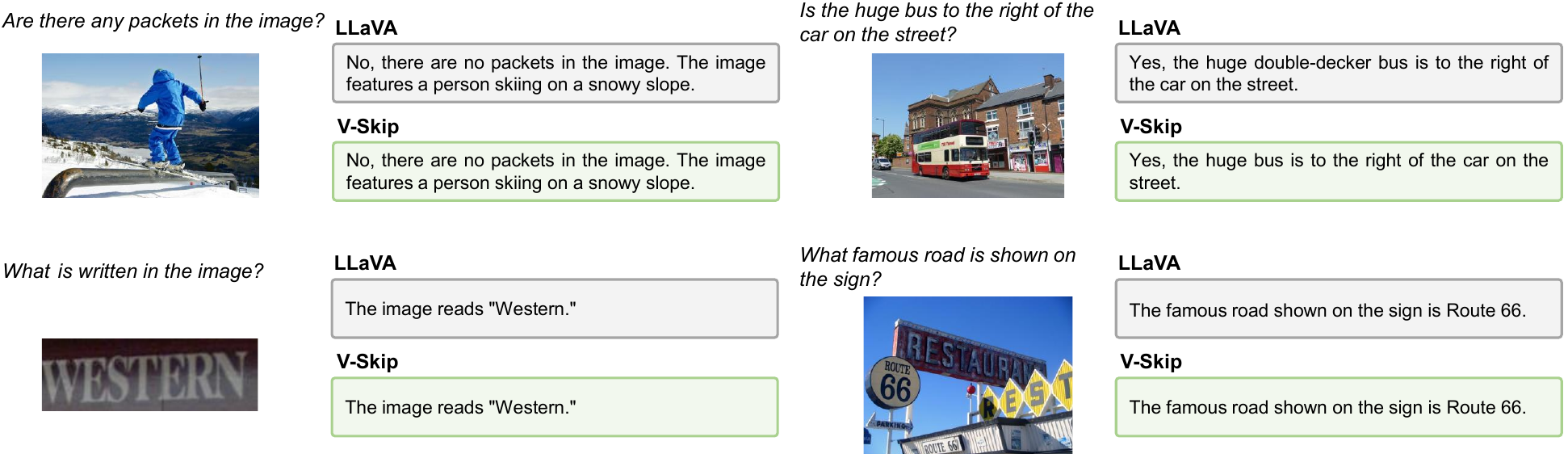}
  \caption{Qualitative comparison of V-Skip and the LLaVA-1.5-7B across activity recognition, spatial reasoning, and fine-grained OCR tasks.
  }
  \label{fig:Qualitative_experiment}
\end{figure}

We compare V-Skip with the LLaVA to evaluate qualitative performance across diverse visual reasoning tasks. As illustrated in \cref{fig:Qualitative_experiment}, V-Skip maintains robust capabilities in object recognition, spatial reasoning, and fine-grained OCR. In the skiing scenario, it accurately identifies the activity while avoiding irrelevant object hallucinations, preserving visual integrity. In urban environments, it correctly perceives the spatial positioning of a double-decker bus relative to other vehicles. Notably, V-Skip excels in text-centric tasks, precisely identifying signs such as \textit{Western} and \textit{Route 66}. These results confirm that by decoupling spatial interaction from semantic evolution through FFN preservation, V-Skip successfully skips approximately 60\% of attention saturation modules while maintaining the original reasoning depth and perceptual capability of the dense model.

\section{Conclusions}

In this paper, we investigate the layer-wise visual processing mechanisms within MLLMs and identify the phenomenon of Visual Attention Saturation. We demonstrate that visual saturation dynamically varies across layers, showing that while spatial mixing completes early, semantic alignment via FFNs remains essential throughout the network. Based on this observation, we propose V-Skip, which is a training-free structural optimization paradigm that introduces block-wise sparsity through a robust few-shot calibration. By strategically decoupling interactions, V-Skip successfully skips approximately 60\% of attention saturation modules while preserving the original perceptual capability of the model. Extensive experiments confirm that our approach maintains 98.42\% to 100.31\% performance retention on the LLaVA series and a robust 94.16\% on Qwen, while effectively mitigating object hallucinations. Ultimately, this work advocates for a paradigm shift from aggressive token removal to refined structural decoupling, offering a more robust path for efficient and trustworthy MLLM inference.

\bibliographystyle{unsrt}
\bibliography{references}

\newpage
\appendix
\section{More Implementation Details}
\subsection{Detailed Experimental Environment.}

To ensure complete reproducibility, all inferences and evaluations in this study were conducted on a server running Ubuntu 20.04.6 LTS. The system hardware configuration consists of an Intel(R) Xeon(R) Gold 5318Y CPU @ 2.10GHz, 256GB of system RAM, and NVIDIA GeForce RTX 3090 GPUs, each equipped with 24GB of VRAM.

The software environment is built upon Python 3.10 and the NVIDIA CUDA Toolkit 12.8. For core deep learning implementations and model loading, we utilize PyTorch 2.1.2 paired with torchvision 0.16.2, alongside transformers 4.37.2 and accelerate 1.12.0. Furthermore, to guarantee standardized, fair, and reproducible testing across all reported multimodal benchmarks, such as MME, MMBench, and OCRBench, we strictly employ the official lmms-eval framework and follow the default evaluation protocol, specifically utilizing version 0.3.0.

\section{Additional Compute-Matched Analysis}
\label{sec:app_compute_matched}

We further examine whether V-Skip's accuracy-efficiency trade-off comes from a larger compute budget or from its structural choice to bypass only visual self-attention. The results in \cref{tab:app_isoflops,tab:app_tokenprune,tab:app_ffnskip} compare methods under matched FLOPs, expanded token-pruning baselines, and increasingly aggressive FFN-Skip settings. These analyses consistently indicate that V-Skip's advantage comes from preserving the FFN pathway while removing redundant visual attention computation.

\begin{figure*}[h]
    \centering
    \begin{subfigure}{\linewidth}
        \centering
        \includegraphics[width=0.75\linewidth, trim=0pt 25pt 0pt 0pt, clip]{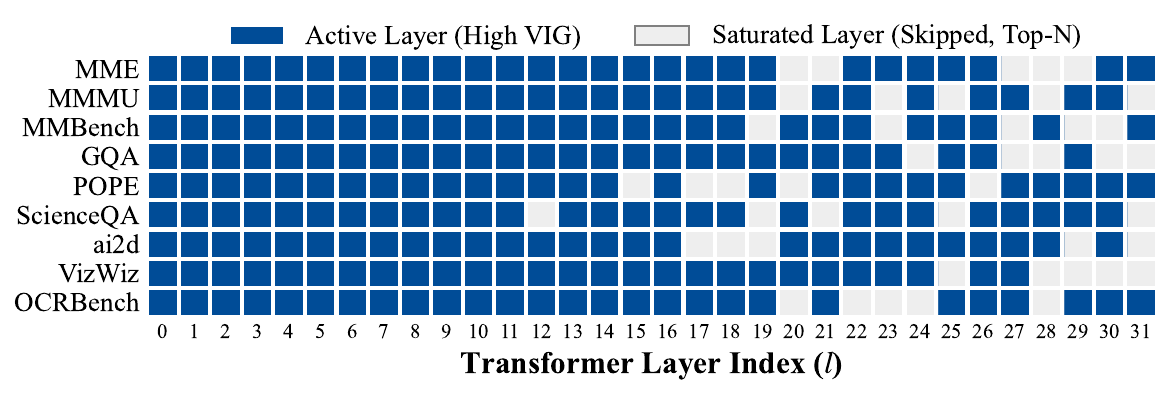}
        \vspace{-2pt}
        \caption{Sparsity Budget $\text{N}=5$}
    \end{subfigure}
    \vspace{-10pt} 
    \\
    
    \begin{subfigure}{\linewidth}
        \centering
        \includegraphics[width=0.75\linewidth, trim=0pt 25pt 0pt 25pt, clip]{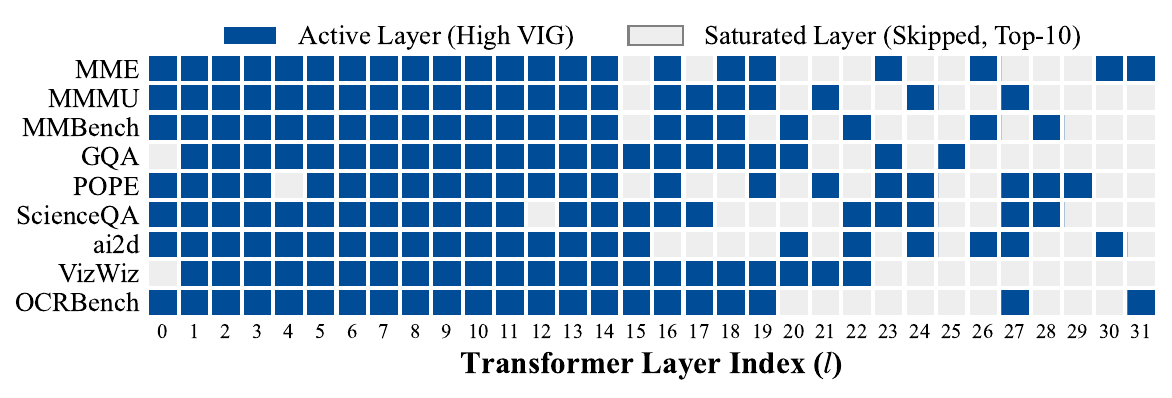}
        \vspace{-2pt}
        \caption{Sparsity Budget $\text{N}=10$}
    \end{subfigure}
    \vspace{-10pt}
    \\
    
    \begin{subfigure}{\linewidth}
        \centering
        \includegraphics[width=0.75\linewidth, trim=0pt 25pt 0pt 25pt, clip]{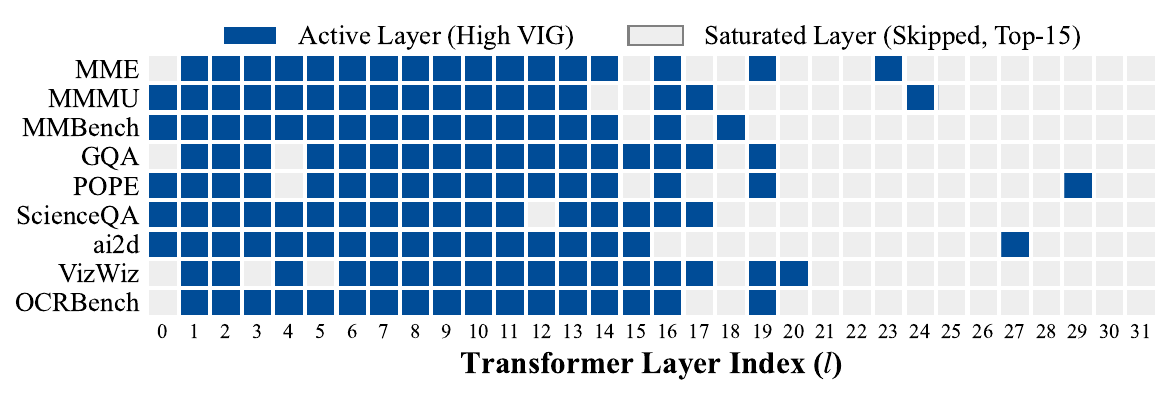}
        \vspace{-2pt}
        \caption{Sparsity Budget $\text{N}=15$}
    \end{subfigure}
    \vspace{-10pt}
    \\
    
    \begin{subfigure}{\linewidth}
        \centering
        \includegraphics[width=0.75\linewidth, trim=0pt 0pt 0pt 25pt, clip]{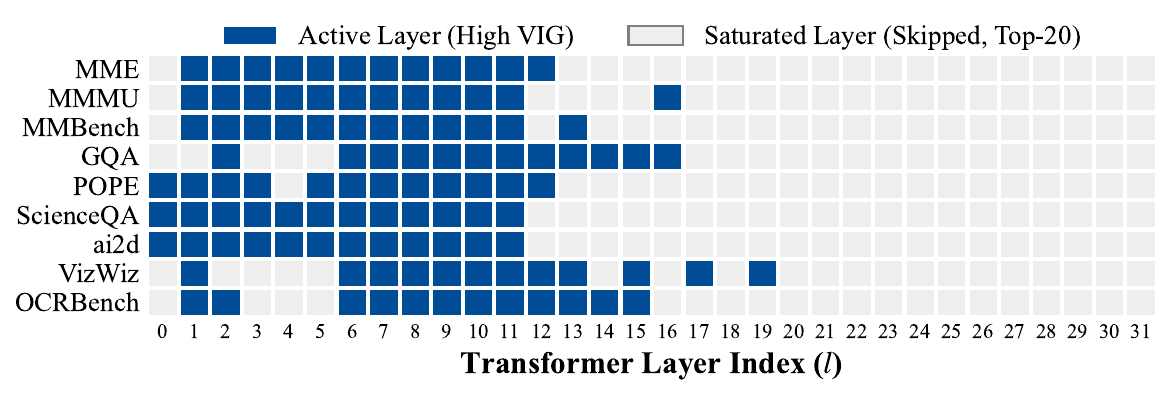}
        \vspace{-2pt}
        \caption{Sparsity Budget $\text{N}=20$}
    \end{subfigure}
    \vspace{-10pt}

    \caption{Evolution of task-specific skipped layers across different sparsity budgets ($N \in \{5, 10, 15, 20\}$). Blue cells indicate active layers (High VIG), while gray cells represent saturated layers dynamically selected for skipping. As the budget increases, the skipping pattern consistently propagates from the deepest layers toward the middle, strictly preserving the early spatial aggregation layers.}
    \label{fig:budget_evolution}
        \vspace{-1em}
\end{figure*}


\begin{table}[h]
\centering
\caption{Iso-FLOPs comparison on LLaVA-1.5-7B at 7.69 TFLOPs.}
\label{tab:app_isoflops}
\setlength{\tabcolsep}{7pt}
\renewcommand{\arraystretch}{1.0}
\begin{tabular}{lccc}
\toprule
Method & TFLOPs & \#L & Avg. \\
\midrule
Dense & 8.54 & 0 & 100.00 \\
ShortV ($N{=}4$) & 7.69 & 4 & 99.92 \\
VTW ($K{=}28$) & 7.69 & 4 & 99.91 \\
\textbf{V-Skip ($N{=}19$)} & \textbf{7.69} & \textbf{19} & \textbf{100.31} \\
\bottomrule
\end{tabular}
\end{table}

\begin{table}[h]
\centering
\caption{Expanded token-pruning baselines under representative configurations on LLaVA-1.5-7B.}
\label{tab:app_tokenprune}
\setlength{\tabcolsep}{10pt}
\renewcommand{\arraystretch}{1.0}
\begin{tabular}{lc}
\toprule
Method & Avg. \\
\midrule
Dense & 100.00 \\
ShortV & 97.73 \\
VTW & 88.71 \\
PyramidDrop & 99.76 \\
SparseVLM & 99.88 \\
\textbf{V-Skip} & \textbf{100.31} \\
\bottomrule
\end{tabular}
\end{table}

\begin{table}[h]
\centering
\caption{V-Skip vs. FFN-Skip across larger sparsity budgets on LLaVA-1.5-7B.}
\label{tab:app_ffnskip}
\setlength{\tabcolsep}{10pt}
\renewcommand{\arraystretch}{1.0}
\begin{tabular}{cccc}
\toprule
$N$ & V-Skip & FFN-Skip & Gap \\
\midrule
19 & \textbf{100.31} & 99.16 & $+1.15$ \\
28 & \textbf{98.68} & 92.14 & $+6.54$ \\
32 & \textbf{91.75} & 71.59 & $+20.16$ \\
\bottomrule
\end{tabular}
\end{table}

\noindent\textbf{Iso-FLOPs comparison.}
When ShortV and VTW are tuned to match V-Skip's 7.69 TFLOPs, V-Skip skips 19 layers while ShortV and VTW skip only four layers each, yet V-Skip still achieves the highest average retention. This suggests that the gain is not simply a matter of using a more permissive compute budget; rather, the attention-only bypass enables a substantially denser sparsity pattern without disrupting the semantic transformation path.

\noindent\textbf{Expanded token-pruning baselines.}
Compared with representative token-pruning methods, including PyramidDrop and SparseVLM, V-Skip is the only method in this comparison that matches or slightly exceeds the dense baseline average. The contrast is especially pronounced on grounding-sensitive settings, where destructive token withdrawal can remove fine-grained visual evidence. For example, OCRBench retention is 31.6 for V-Skip compared with 5.1 for VTW in the main comparison.

\noindent\textbf{FFN preservation under larger sparsity budgets.}
The gap between V-Skip and FFN-Skip is modest at $N{=}19$ because this is a conservative sparsity regime. As the sparsity budget increases, the gap widens to 6.54 points at $N{=}28$ and 20.16 points at $N{=}32$. At $N{=}32$, FFN-Skip drops sharply on GQA (61.9$\to$39.6) and POPE (85.1$\to$61.5), while V-Skip retains 54.5 and 78.7. These results support the view that FFNs remain important for cross-modal projection even when deep visual self-attention is redundant.

\noindent\textbf{Latency and deployment note.}
The layer-aligned latency comparison in the main paper isolates what is bypassed inside each layer: V-Skip preserves FFNs, while ShortV bypasses entire blocks. V-Skip therefore gives a more moderate standalone latency reduction of approximately 8\%, but it preserves accuracy more reliably and can be combined with token pruning for stronger acceleration, as shown by FastV+V-Skip. In a deployment-oriented throughput measurement, V-Skip improves throughput from 14.4 to 14.8 while keeping the KV-cache footprint unchanged, consistent with its role as a prefill-compute accelerator.

\section{Budget Evolution and Task-Specific Sparsity}
\label{sec:budget_evolution}

To comprehensively understand the dynamics of our task-aware calibration mechanism, we visualize the evolution of active and skipped vision attention across increasing sparsity budgets ($\text{N} \in \{5, 10, 15, 20\}$) for nine representative benchmarks \cref{fig:budget_evolution}.
This visualization yields two critical architectural insights regarding how MLLMs process visual information under varying Visual Information Gain constraints.

\subsection{Dynamic Deep-to-Shallow Progression.} 
Analyzing the evolution across different budgets demonstrates that our few-shot calibration is a highly dynamic ranking process. Under a conservative budget ($\text{N}=5$), the mechanism universally prioritizes the deepest layers (e.g., layers 20--31), confirming that visual attention redundancy initially emerges at the terminal stages of semantic alignments. As the sparsity budget expands to $\text{N}=20$, the skipped saturated layers propagate a trend toward the middle or early layers. Crucially, the early transformer blocks (layer 1--11) remain overwhelmingly active across all configurations, acting as a foundation for spatial feature interaction. This consistent evolution trajectory proves that VIG metric provides a way for dynamically scaling block-wise redundancy rather than applying a fixed, predefined pruning template.
\begin{center}
\begin{tcolorbox}[colback=gray!10,colframe=black,width=\linewidth, arc=0.5mm,auto outer arc,boxrule=0.5pt]
    \textit{\textbf{Takeaway:}} Deep layers are identified as the primary source of visual attention saturation, whereas early layers act as an indispensable structural cornerstone that V-Skip autonomously protects.
\end{tcolorbox}
\end{center}

\subsection{Task-Driven Routing Consistency.}
Observing the permutation of skipped visual attention saturation layers uncovers a correlation between task characteristics and the evolution of visual saturation.
Benchmarks with similar reasoning and visual perception demand naturally cluster into distinct evolutionary patterns at higher sparsity budgets (e.g., $\text{N}=20$). General multimodal evaluation, such as MME and MMBench, exhibit remarkably similar structures aggressively bypassing deep blocks and dropping the initial projection layer. In contrast, structure sensitive tasks like OCRBench share a fundamentally different pattern. They strictly preserve layer 0 and selectively retain specific intermediate layers to interact with visual cues.
\begin{center}
\begin{tcolorbox}[colback=gray!10,colframe=black,width=\linewidth, arc=0.5mm,auto outer arc,boxrule=0.5pt]
    \textit{\textbf{Takeaway:}} V-Skip acts as a domain-aware router. It empirically proves that a ``one-fits-all'' pruning mask is suboptimal, as fine-grained reasoning inherently requires distinct semantic pathways compared to general perception.
\end{tcolorbox}
\end{center}

\begin{table}[t]
    \centering
        \caption{Calibration time overhead across different multimodal benchmarks. The recorded time represents the total duration (mm:ss) required to compute the VIG and identify the optimal block-wise skipping strategy using our 20-shot subset on a single NVIDIA 3090 GPU.}
    \setlength{\tabcolsep}{3.5pt}
    \footnotesize

    \resizebox{0.3\linewidth}{!}{
    \begin{tabular}{l c}
        \hline
        \textbf{Dataset} & \textbf{Calibration Time} \\
        \hline
        \rowcolor{gray!15} 
        MME & 01:55 \\
        MMMU & 02:06 \\
        \rowcolor{gray!15} 
        MMBench & 02:15 \\
        GQA & 01:54 \\
        \rowcolor{gray!15} 
        POPE & 01:42 \\
        SQA & 02:10 \\
        \rowcolor{gray!15} 
        AI2D & 02:03 \\
        VizWiz & 02:03 \\
        \rowcolor{gray!15} 
        OCRBench & 01:39 \\
        \hline
    \end{tabular}}
    \label{tab:calibration_time}
\end{table}

\subsection{Calibration Efficiency and Overhead.}
\label{sec:calibration_time}

It is crucial that task-specific calibration pre-computation does not introduce prohibitive delays while guaranteeing optimal routing.
To quantify this overhead, we measure the wall-clock time required to compute the layer-wise VIG scores and extract the Top-N skipping configuration. All calibrations utilize our default 20-shot sampling strategy and are executed on a single NVIDIA 3090 GPU.
As reported in~\cref{tab:calibration_time}, the entire calibration process is remarkably lightweight. For the majority of downstream tasks, it completes in approximately two minutes (e.g., 01:39 for OCRBench and 01:55 for MME). 
Given that this is a strict one-time offline cost incurred prior to inference, the overhead is practically negligible. When weighed against the substantial inference acceleration and robust performance retention that V-Skip delivers throughout the model's deployment lifecycle, this minimal pre-computation firmly establishes our method as a practical, plug-and-play solution.

\section{Further Discussion on Functional Saturation}
\label{sec:app_functional_saturation}

Both VIG and ShortV use KL divergence, but they perturb different computational targets and therefore measure different phenomena. ShortV perturbs entire transformer blocks, jointly removing attention and FFN computation. This produces a single whole-block sensitivity signal that conflates spatial interaction and semantic projection. In contrast, VIG isolates the visual-to-visual self-attention pathway while preserving FFN transformations. This targeted perturbation exposes a modality-specific functional asymmetry: visual self-attention can saturate earlier than the FFN pathway, while FFNs remain important for aligning visual tokens with the evolving language-model semantic space.

This distinction also leads to different interventions. ShortV freezes both attention and FFN updates in skipped layers, whereas V-Skip bypasses only saturated visual attention and keeps FFNs active for all tokens. The comparison in \cref{tab:app_ffnskip} shows that bypassing FFNs becomes increasingly harmful as the sparsity budget grows, which would be hidden by a whole-block scoring rule. VIG is therefore best understood as a functional sensitivity measure: it quantifies how much the final predictive distribution changes when visual self-attention at a layer is bypassed. It does not directly claim to measure internal representation saturation; instead, it identifies whether a layer's visual attention computation is functionally necessary for downstream predictions.

The LLaVA-NeXT-7B MME result further illustrates this functional view. V-Skip drops from 1846.3 to 1778.6 on MME, a 3.7\% reduction and the largest single-benchmark drop in the reported experiments. This indicates that high-resolution settings can retain a stronger dependence on deep-layer visual interaction for some tasks, even though the average retention across nine benchmarks remains 98.42\%.

Finally, lmms-eval is deterministic under the reported protocol, so the observed gaps primarily reflect structural differences rather than sampling noise. For this reason, claims around V-Skip should be interpreted as evidence that attention/FFN decoupling preserves visual grounding and avoids exacerbating hallucination under the evaluated settings, rather than as a universal hallucination-mitigation guarantee.

\section{Exploring Universal Calibration for V-Skip}
\label{sec:universal_skip}

Our proposed V-Skip employs task-specific calibration, identifying the optimal saturated layers for each individual dataset to maximize performance retention. To further investigate the transferability and robustness of the visual attention saturation phenomenon, we conduct an additional ablation study evaluating a "universal" layer skipping strategy. Specifically, we introduce \textbf{V-Skip*}, which derives its layer-skipping configuration purely calibrated on the MME benchmark and applies this exact same strategy globally across all other evaluation datasets without any task-specific tuning.

As presented in \cref{tab:universal_skip}, the universally calibrated V-Skip* demonstrates strong generalization, achieving an impressive average performance retention of 99.55\% across all benchmarks. It consistently outperforms other training-free acceleration baselines across various benchmarks, significantly surpassing ShortV (97.73\%) and VTW (88.71\%). This clearly highlights the inherent architectural superiority of our attention-bypassing and FFN-preserving paradigm compared to destructive token dropping or complete layer freezing.

However, we also observe a distinct performance gap between the universal V-Skip* (99.55\%) and our default, task-specifically calibrated V-Skip (100.31\%). The degradation is more noticeable in tasks requiring fine-grained visual grounding, such as OCRBench (dropping from 31.6 to 30.7) and MMBench (dropping from 65.0 to 64.4). While a universal configuration is highly competitive and adaptable, diverse multimodal reasoning tasks inherently demand different depths of spatial attention. Therefore, to achieve optimal performance and fully preserve the model's reasoning integrity, a fine-grained calibration tailored to the specific data distribution of the target task remains essential. Nonetheless, the strong baseline established by V-Skip* demonstrates its viability as a highly practical, out-of-the-box acceleration solution for real-world deployments where task-agnostic efficiency is prioritized.

\begin{table}[!t]
    \centering
    \setlength{\tabcolsep}{3.5pt}
    \footnotesize
    \caption{Performance comparison of universal layer skipping. V-Skip* applies the MME-calibrated layer skipping strategy globally to all other benchmarks, whereas V-Skip uses task-specific calibration.}
    \label{tab:universal_skip}
    \resizebox{\linewidth}{!}{
        \begin{tabular}{l | c c c c c c c c c | c}
            \hline
            \textbf{Method} & \textbf{MME} & \textbf{MMMU} & \textbf{MMB} & \textbf{GQA} & \textbf{POPE} & \textbf{SQA} & \textbf{AI2D} & \textbf{VizWiz} & \textbf{OCRB} & \makecell{\textbf{Avg.}} \\
            \hline
            \rowcolor{gray!15} 
            LLaVA-1.5-7B       & 1866.1 & 36.7 & 64.1 & 61.9 & 85.1 & 69.6 & 55.2 & 54.3 & 31.3 & 100.00\% \\
            ShortV (N=19)      & 1842.2 & 36.2 & 64.8 & 60.8 & 84.4 & 68.3 & 54.2 & 50.6 & 29.5 & 97.73\% \\
            VTW (K=16)         & 1857.6 & 36.4 & 64.0 & 55.1 & 85.3 & 69.6 & 55.4 & 51.0 & 5.1  & 88.71\% \\
            \rowcolor{green!10} 
            V-Skip* (N=19)     & 1853.3 & 36.6 & 64.4 & 61.4 & 85.2 & 69.2 & 55.4 & 53.9 & 30.7 & \textbf{99.55\%} \\
            \rowcolor{green!15} 
            V-Skip (N=19)      & 1853.3 & 36.9 & 65.0 & 61.8 & 85.5 & 69.3 & 55.4 & 54.5 & 31.6 & \textbf{100.31\%} \\
            \hline
        \end{tabular}
    }
\end{table}

\section{Qwen2.5-VL Re-Calibration Analysis}
\label{sec:app_qwen_recalibration}

The Qwen2.5-VL results show that the optimal sparsity budget is architecture-dependent. The original $N{=}7$ setting follows the approximate skip ratio used for LLaVA-style architectures, but it is relatively aggressive for Qwen2.5-VL. Unlike LLaVA, whose VIG profile shows a clearer deep-layer saturation plateau, Qwen2.5-VL exhibits persistently higher VIG across depth, likely because window attention and mRoPE redistribute visual redundancy. Re-calibrating the sparsity budget accordingly substantially recovers performance.

\begin{table}[t]
\centering
\caption{Qwen2.5-VL-7B re-calibration. Reducing the sparsity budget from the submitted $N{=}7$ configuration to $N{=}4$ better matches Qwen2.5-VL's VIG profile and recovers average retention to 98.72\%.}
\label{tab:app_qwenrecal}
\setlength{\tabcolsep}{5pt}
\renewcommand{\arraystretch}{1.0}
\begin{tabular}{lcccc}
\toprule
Config & MMMU & GQA & VizWiz & Avg. Retention \\
\midrule
Dense & 50.4 & 60.5 & 70.2 & 100.00\% \\
\textbf{$N{=}4$} & \textbf{46.7} & \textbf{58.3} & \textbf{69.2} & \textbf{98.72\%} \\
$N{=}5$ & 46.9 & 54.7 & 67.1 & 96.67\% \\
$N{=}6$ & 46.0 & 52.2 & 67.3 & 95.27\% \\
$N{=}7$ & 43.3 & 51.3 & 66.4 & 94.16\% \\
$N{=}8$ & 41.9 & 48.0 & 65.8 & 92.07\% \\
\bottomrule
\end{tabular}
\end{table}

Moving from $N{=}7$ to $N{=}4$ improves GQA by 7.0 points and MMMU by 3.4 points, raising Qwen2.5-VL's overall retention from 94.16\% to 98.72\%. This aligns Qwen2.5-VL with the near-lossless regime observed for LLaVA-NeXT and supports the use of architecture-specific calibration rather than a fixed global skip ratio.

\section{Limitations and Future Work}

While V-Skip demonstrates significant efficiency gains and maintains high performance across various multimodal benchmarks, we acknowledge certain limitations. A core advantage of V-Skip is its training-free, plug-and-play nature. However, because the original model weights were optimized under dense attention assumptions, bypassing these paths at inference time represents a structural shift from the initial pre-training phase. Consequently, our current approach relies on a lightweight few-shot calibration mechanism to dynamically identify the task-optimal sparsity path, accounting for the varying depths of spatial reasoning required by diverse downstream tasks.

To achieve a more universal, task-agnostic generalization---where the network natively learns to route or bypass visual computation without prior empirical calibration---future work will focus on incorporating the V-Skip mechanism directly into the model's training pipeline. Integrating this block-wise skipping during multimodal pre-training or instruction tuning would allow the model to natively internalize the Visual Attention Saturation phenomenon. This co-design of architecture and training could eliminate the reliance on task-specific calibration, potentially unlocking even greater synergies between performance and computational efficiency.

\section{Additional Visualizations}
\label{sec:additional_visualizations}

To further illustrate the practical advantages of V-Skip and provide a deeper understanding of how different acceleration paradigms impact multimodal reasoning, we provide extended qualitative comparisons and a Qwen2.5-VL VIG visualization. \cref{fig:qualitative_gqa,fig:qualitative_ocr} present challenging cases from the GQA and OCRBench datasets, respectively. \cref{fig:app_qwenvig} visualizes the Qwen2.5-VL VIG profile discussed in \cref{sec:app_qwen_recalibration}.

\subsection{Mitigating Spatial Misalignment and Hallucinations}
\label{sec:app_visual_gqa}
\cref{fig:qualitative_gqa} highlights tasks requiring precise spatial awareness and fine-grained object grounding within complex indoor and outdoor scenes. Aggressive token reduction methods like VTW tend to lose critical localized features, resulting in generic or less precise answers, such as predicting ``Computer'' instead of ``Keyboard'', or ``Man'' instead of ``Policeman''. Conversely, layer-skipping methods like ShortV, which completely bypass the FFNs, suffer from severe feature misalignment. This semantic disruption leads to obvious spatial errors and object hallucinations. V-Skip preserves FFN transformations for continuous semantic evolution, enabling accurate grounding of fine-grained objects and complex spatial relations.

\subsection{Preserving Structural Integrity for OCR Tasks}
\label{sec:app_visual_ocr}
Recognizing text in the wild requires high-resolution preservation and structural integrity. As shown in \cref{fig:qualitative_ocr}, VTW tends to destroy textual feature sequences and can produce generic placeholders such as ``Words'' or ``text''. ShortV lacks the FFN projections needed to map complex visual text patterns into the language-model vocabulary space, producing spelling hallucinations. By bypassing only redundant spatial mixing while keeping point-wise semantic projections active, V-Skip maintains strong perceptual fidelity on cursive fonts, stylized text, and numerical sequences.

\begin{figure}[!t]
    \centering
    \includegraphics[width=0.85\linewidth]{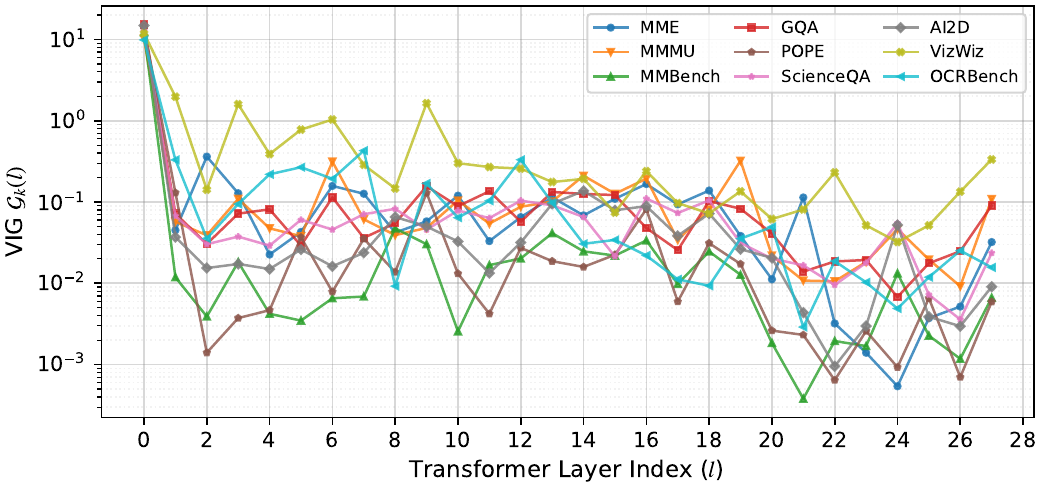}
    \caption{Qwen2.5-VL VIG profile. Compared with the clearer deep-layer saturation observed in LLaVA-style models, Qwen2.5-VL maintains higher VIG across more layers, motivating the architecture-specific recalibration in \cref{tab:app_qwenrecal}.}
    \label{fig:app_qwenvig}
\end{figure}

\subsection{Qwen2.5-VL VIG Profile}
\label{sec:app_visual_qwen}
\cref{fig:app_qwenvig} shows that Qwen2.5-VL does not exhibit the same clear deep-layer saturation plateau as LLaVA-style architectures. Its VIG remains relatively high across a broader portion of the depth, explaining why a smaller sparsity budget is more appropriate for this architecture.

\begin{figure}[!t]
    \centering
    \includegraphics[width=0.65\linewidth]{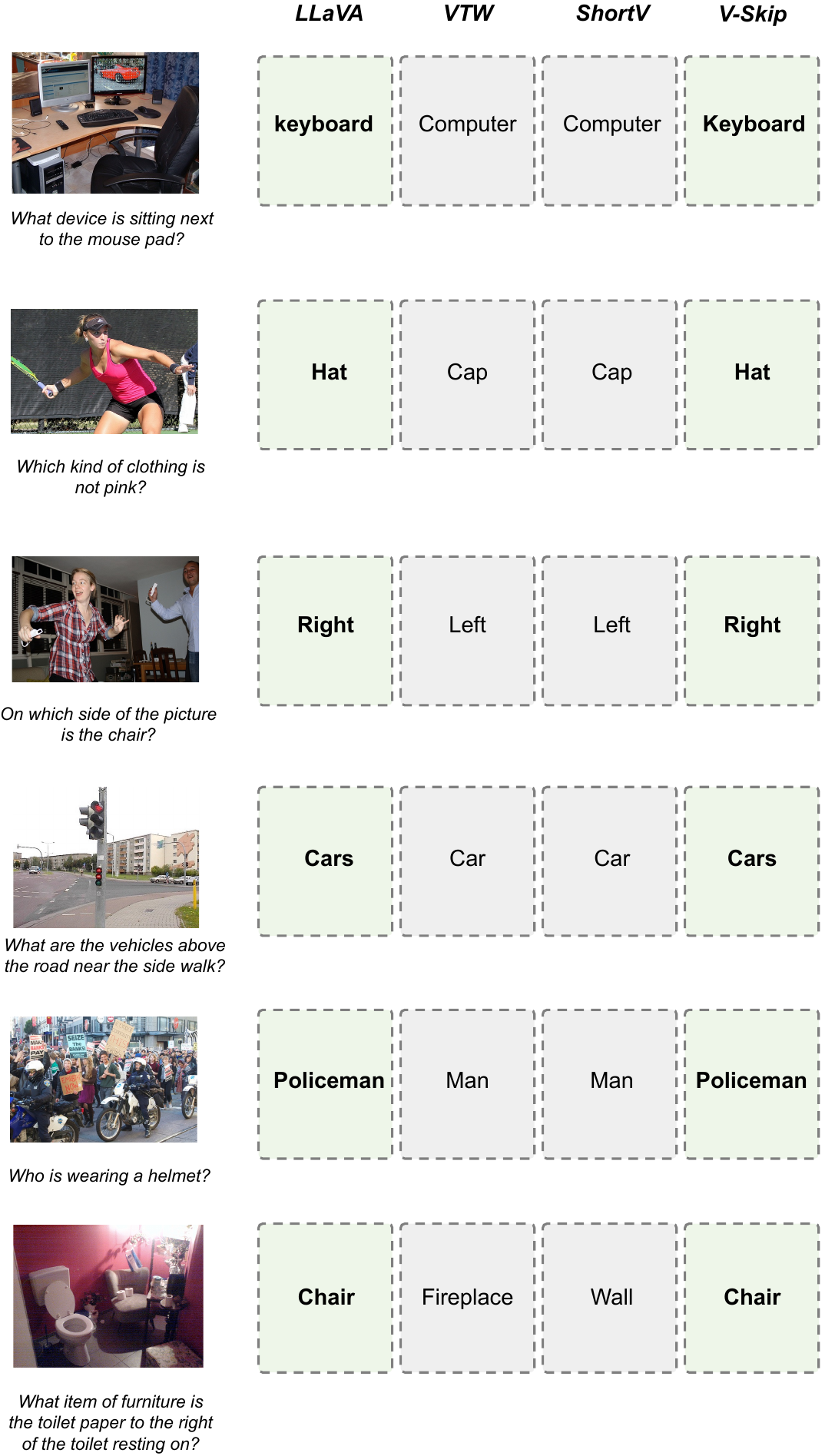}
    \caption{Qualitative comparison on spatial reasoning and object grounding. V-Skip preserves fine-grained visual details and complex spatial relations, while destructive token withdrawal and full-layer skipping can lead to generic outputs, spatial reasoning errors, and object hallucinations.}
    \label{fig:qualitative_gqa}
\end{figure}

\begin{figure}[t]
    \centering
    \includegraphics[width=0.75\linewidth]{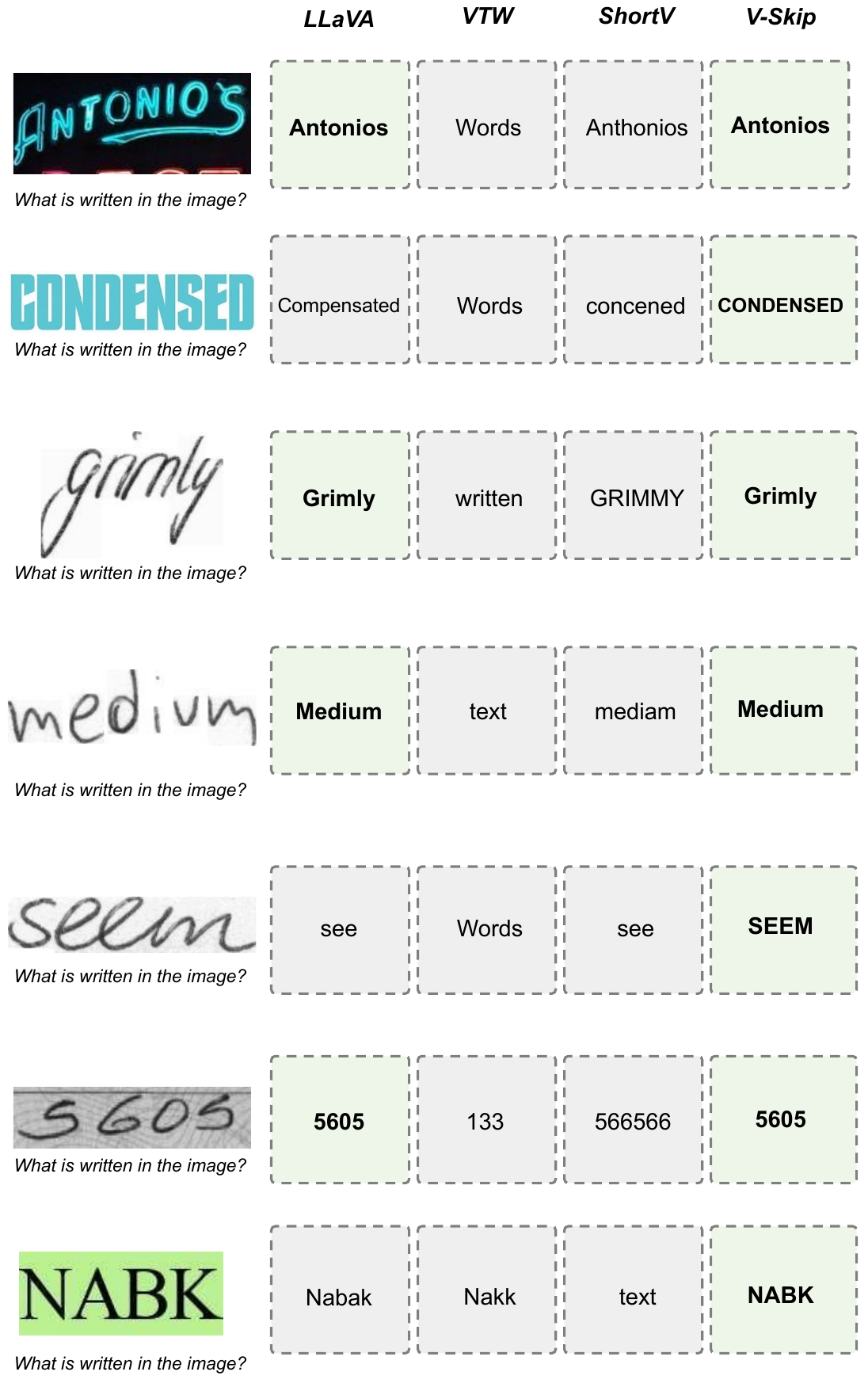}
    \caption{Qualitative comparison on OCR samples. Recognizing stylized, cursive, and handwritten text requires structural integrity. V-Skip preserves fine-grained textual features more reliably than destructive token withdrawal or full-layer skipping.}
    \label{fig:qualitative_ocr}
\end{figure}

\end{document}